\documentclass{article}
\usepackage{microtype}
\usepackage{graphicx}
\usepackage{subfigure}
\usepackage{booktabs} % for professional tables

\usepackage{hyperref}

\usepackage[accepted]{icml2024}
\usepackage{amsmath,amsfonts,bm}

\def\eqref#1{equation~\ref{#1}}
\def\1{\bm{1}}

\DeclareMathAlphabet{\mathsfit}{\encodingdefault}{\sfdefault}{m}{sl}
\SetMathAlphabet{\mathsfit}{bold}{\encodingdefault}{\sfdefault}{bx}{n}

\definecolor{MyDarkRed}{rgb}{0.8,0.02,0.02}
\definecolor{MyDarkBlue}{rgb}{0.02,0.02,0.8}
\definecolor{MyDarkGreen}{rgb}{0.1,0.8,0.1}

\newcommand{\dataset}{ContPhy\xspace}
\newcommand{\datasetFull}{Continuum Physical Dataset}
\newcommand{\model}{ContPRO\xspace}

\makeatletter
\DeclareRobustCommand\onedot{\futurelet\@let@token\@onedot}
\def\@onedot{\ifx\@let@token.\else.\null\fi\xspace}
\def\eg{\emph{e.g}\onedot} 
\def\ie{i.e\onedot} 
\usepackage{amsmath}
\usepackage{amssymb}
\usepackage{mathtools}
\usepackage{amsthm}

\usepackage{xspace}
\usepackage{multirow}
\usepackage{comment}
\usepackage{graphicx}
\usepackage{wrapfig}
\usepackage{makecell}
\usepackage{multicol}
\usepackage{multirow}

\usepackage{listings}

\definecolor{codegreen}{rgb}{0,0.6,0}
\definecolor{codegray}{rgb}{0.5,0.5,0.5}
\definecolor{codepurple}{rgb}{0.58,0,0.82}
\definecolor{backcolour}{rgb}{0.97,0.99,0.97}

\lstdefinestyle{mystyle}{
    backgroundcolor=\color{backcolour},   
    commentstyle=\color{codegreen},
    keywordstyle=\color{magenta},
    numberstyle=\tiny\color{codegray},
    stringstyle=\color{codepurple},
    basicstyle=\ttfamily\scriptsize,
    breakatwhitespace=false,         
    breaklines=true,                 
    captionpos=b,                    
    keepspaces=true,                 
    numbers=left,                    
    numbersep=5pt,                  
    showspaces=false,                
    showstringspaces=false,
    showtabs=false,                  
    tabsize=2,
    showlines=true
}

\usepackage[capitalize,noabbrev]{cleveref}

\newlength\savewidth
\newcommand\shline{\noalign{\global\savewidth\arrayrulewidth
  \global\arrayrulewidth 1pt}\hline\noalign{\global\arrayrulewidth\savewidth}}

\theoremstyle{plain}

\theoremstyle{definition}

\theoremstyle{remark}

\usepackage[textsize=tiny]{todonotes}

\icmltitlerunning{ContPhy for Continuum Physical Reasoning}

\begin{document}

\twocolumn[
\icmltitle{ContPhy: Continuum Physical Concept Learning and Reasoning from Videos}



\icmlsetsymbol{equal}{*}

\begin{icmlauthorlist}
\icmlauthor{Zhicheng Zheng}{equal,thu}
\icmlauthor{Xin Yan}{equal,whu}
\icmlauthor{Zhenfang Chen}{equal,ibm}
\icmlauthor{Jingzhou Wang}{thu}
\icmlauthor{Qin Zhi Eddie Lim}{mit}
\icmlauthor{Joshua B. Tenenbaum}{mit}
\icmlauthor{Chuang Gan}{ibm,umass}
\end{icmlauthorlist}

\icmlaffiliation{thu}{Tsinghua University}
\icmlaffiliation{whu}{Wuhan University}
\icmlaffiliation{ibm}{MIT-IBM Watson AI Lab}
\icmlaffiliation{umass}{UMass Amherst}
\icmlaffiliation{mit}{Massachusetts Institute of Technology}

\icmlcorrespondingauthor{Zhenfang Chen}{chenzhenfang2013@gmail.com}
\icmlcorrespondingauthor{Chuang Gan}{ganchuang1990@gmail.com}

\icmlkeywords{Machine Learning, ICML}

\vskip 0.3in
]



\printAffiliationsAndNotice{\icmlEqualContribution} % otherwise use the standard text.

\begin{abstract}
We introduce the \datasetFull~(\dataset), a novel benchmark for assessing machine physical commonsense. 
\dataset complements existing physical reasoning benchmarks by encompassing the inference of diverse physical properties, such as mass and density, across various scenarios and predicting corresponding dynamics.
We evaluated a range of AI models and found that they still struggle to achieve satisfactory performance on \dataset, which shows that current AI models still lack physical commonsense for the continuum, especially soft-bodies, and illustrates the value of the proposed dataset.
We also introduce an oracle model (\model) that marries the particle-based physical dynamic models with the recent large language models, which enjoy the advantages of both models, precise dynamic predictions, and interpretable reasoning.
\dataset aims to spur progress in perception and reasoning within diverse physical settings, narrowing the divide between human and machine intelligence in understanding the physical world. Project page: {\url{https://physical-reasoning-project.github.io}}. 

\end{abstract}
\section{Introduction}
Humans are capable of comprehending the physical properties of various substances, including rigid objects and soft objects, understanding their dynamic interactions in complex environments, and predicting their corresponding dynamic changes. In fact, this innate ability to understand and reason about the physical world plays a crucial role in shaping our understanding of nature and the development of scientific knowledge~\citep{kill2020mental}.

\begin{figure*}[t]
\begin{center}
\includegraphics[width=0.95\linewidth]{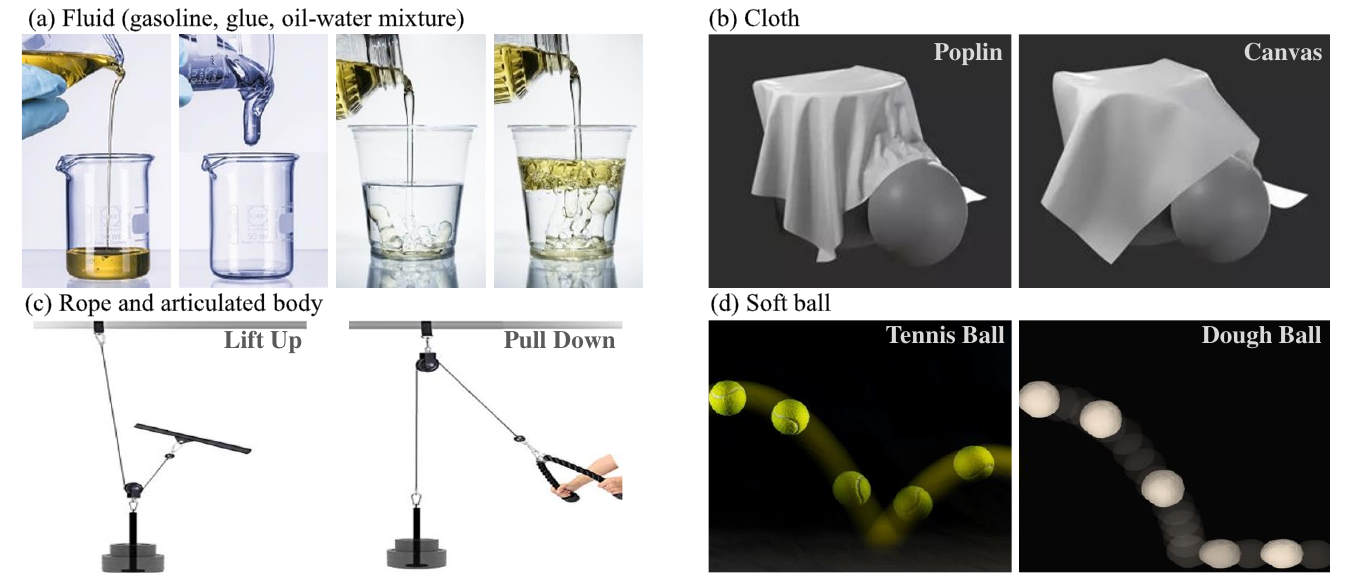}
\vskip 0.1in
\caption{
The motivation is derived from a range of everyday soft materials and their interaction with rigid objects, whose physical behaviors or functions vary by their diverse physical properties. a) Gasoline flows more fluently than glue due to lower viscosity, while oil with lower density tends to float above water. b) Poplin and canvas exhibit surface wrinkles with varying granularity due to their distinct bending compliance. c) The lifting approach requires less force due to the re-distributed tensile forces facilitated by the movable pulley. d) Trajectories of tennis ball and dough ball demonstrate their differing elasticity and plasticity. 
}
\label{fig:teaser}
\end{center}
\end{figure*}

As depicted in Figure~\ref{fig:teaser}, objects like solids and liquids in nature often exhibit different properties, and these objects of different properties couple together to build our complex physical world. As humans, we are able to distinguish objects' physical properties by observing their interactions.We know that the clear liquid in Figure~\ref{fig:teaser}~(a) at the bottom has a higher density than the yellow liquid on the top; we know that the dynamic pulley in Figure~\ref{fig:teaser}~(c) could help us to pull the cargo up more easily. These innate human skills raise an intriguing question: can current AI models have the physical common sense to infer physical properties of the continuum\footnote{Continuum encompasses various bodies like liquids, soft materials (\eg, ropes), rigid bodies and articulated bodies (\eg, pulleys). More details about the physical concept of the continuum can be found in Section~\ref{app:continuum} in the Appendix.} and predict corresponding dynamics?

Recently, a series of benchmarks~\citep{riochet2018intphys,rajani2020esprit,bear1physion}, have been developed to study machine models' effectiveness for physical reasoning. However, there have been limitations that make them non-ideal for the assessment of whether machine models have human-like physical reasoning abilities. 
Firstly, most benchmarks mainly deal with simple visual primitives like spheres, cubes, and collision events of rigid objects only.
It remains doubtful whether the conclusions based on these simple scenes will still hold in more comprehensive visual scenarios with the coupling of soft objects and their interaction with rigid objects.
There have also been benchmarks like Physion~\citep{bear1physion} that were developed to evaluate machine models' physical reasoning abilities in different scenarios.
However, objects in Physion are of the same physical parameters without any variance (\eg solids with the same mass and water with the same density).
Moreover, Physion only requires models to predict whether two objects will come into contact after the observed video ends. It has not incorporated natural language to answer other challenging questions like predicting dynamics in counterfactual scenes and selecting actions to achieve a goal.

To this end, we aim to build a \datasetFull~(\dataset) to thoroughly evaluate and diagnose machine models' physical reasoning performance in comprehensive physical environments. The design of \dataset aims to achieve two goals: 1) covering diverse physical scenarios and 2) supporting comprehensive natural language tasks. 

\begin{figure*}[!t]
\begin{center}
\centerline{\includegraphics[width=.99\linewidth]{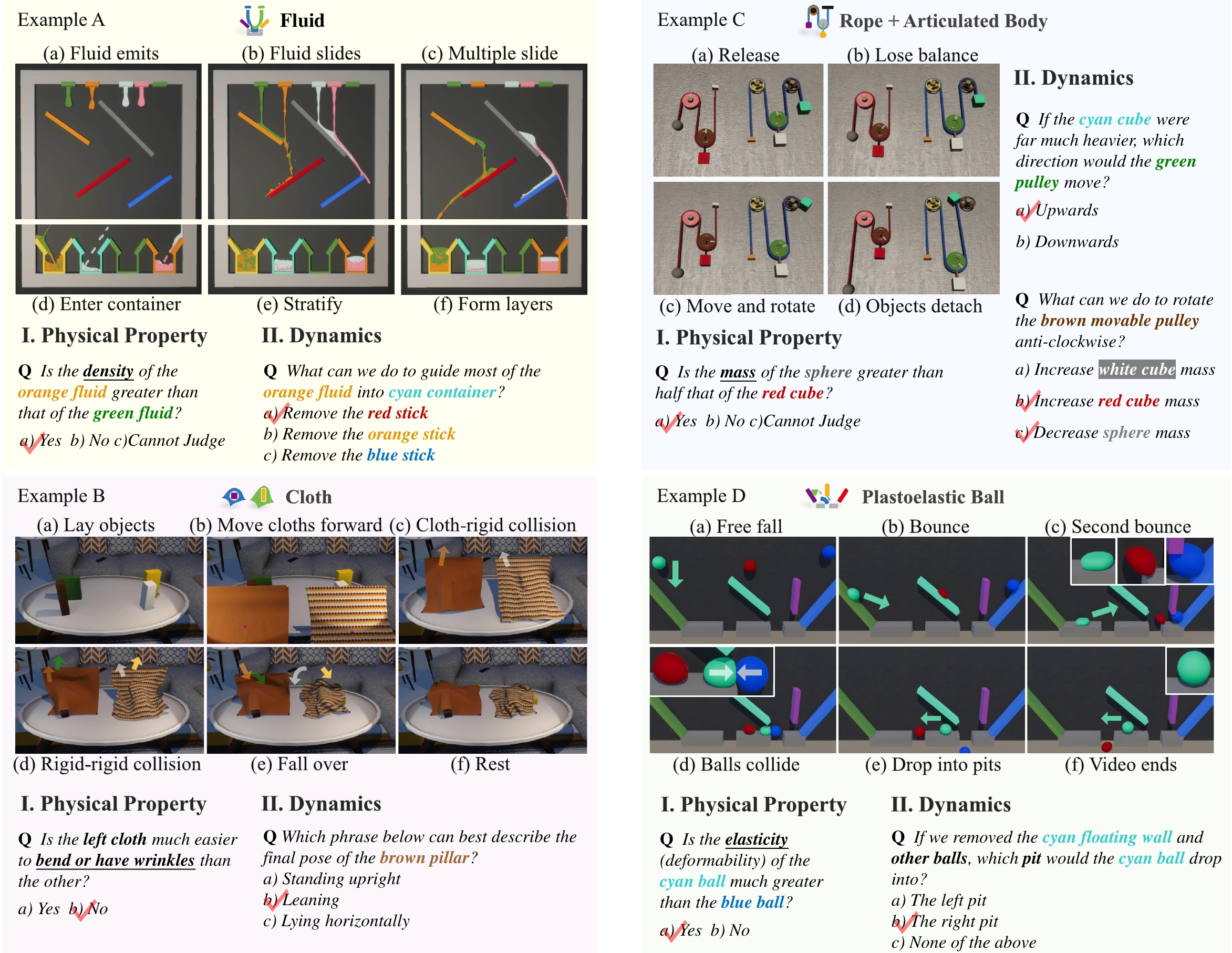}}
\vskip 0.1in
\caption{
  The figure presents samples from the four puzzle blocks of our \datasetFull (\dataset). \dataset offers rendered outputs from the simulation of randomly sampled scenarios, accompanied by their respective question-answer pairs. These pairs span from understanding soft-body physical properties, concepts, and interactions with rigid objects through comparative analysis, to temporal and spatial dynamic predictions, counterfactual considerations, and goal-oriented problem-solving. It aims to provide a comprehensive resource for AI models to interpret the physical world of various deformable bodies. Note that they are best viewed in videos.
}
\label{fig:dataset}
\end{center}
\end{figure*}

To achieve the first goal, we adopt the physical engine~\citep{haas2014history} to simulate diverse videos with dense supervision signals. As shown in Figure~\ref{fig:dataset}, the simulated physical scenes include scenes with the coupling of different liquids, deformable cloths, pulley systems, and elastoplastic balls.
Another goal of the built dataset is to propose diverse physical reasoning tasks in the form of video question answering. We achieve this goal with a carefully designed question engine. The question engine takes the dense simulated video annotation as input and generates different questions based on pre-defined textual templates. Sample questions can be found in Figure~\ref{fig:dataset}. It asks challenging questions such as ``\textit{If the red stick were removed, would most orange fluid flow into the cyan container?}" and ``\textit{Is the mass of the sphere greater than half that of the red cube?}", which requires the model to have a deep understanding of physical scenes and reason about their dynamics.

We also evaluate a series of traditional AI models~\citep{hudson2018compositional,li2022align, le2020hierarchical} and recent multimodal large language models~\citep{team2023gemini,achiam2023gpt} on~\dataset. We found that the performance of these models is far from satisfactory, demonstrating the proposed~\dataset benchmark's value and indicating the necessity of more advanced models with better physical common sense.

To better investigate the characteristics of \dataset and show insights to build stronger physical reasoning models, we introduce an oracle model, ~\model that marries two powerful research ideas, particle-based models~\citep{li2018propagation,sulsky1995application} for dynamic predictions and the recent large language models~\citep{ouyang2022training} for complex language reasoning. While it requires more supervision signals (\ie particle-based representation for the scenes), \model achieves the best overall performance.

To summarize, the contribution of the paper lies in three aspects.
First, we introduce a pioneering benchmark for physical reasoning that encapsulates a wide spectrum of physical properties such as mass, density, elasticity, and deformability. Complementing this, we have developed a meticulously crafted question engine capable of synthesizing a variety of complex physical reasoning queries.
Second, we extensively evaluate the proposed benchmark with multiple machine models to study the characteristics and show insights into physical reasoning model development.
Finally, we develop an oracle model for the benchmark, which combines symbolic representation with particle-based dynamic models for physical understanding and reasoning.

\begin{table*}[t]
\caption{Comparison between \dataset and other physical reasoning benchmarks. \dataset is a dataset that covers a wide variety of tasks including reasoning about the continuum's physical properties, counterfactual dynamics, and goal planning in diverse physical scenarios.}
\label{tab:dataset_comparison}
\vskip 0.1in
    \resizebox{1\linewidth}{!}
    {
    \small
    \setlength{\tabcolsep}{2pt}
	\begin{tabular}{lccccccc}
	\toprule
         \multirow{2}{*}{Dataset}  & Question &  \multirow{2}{*}{Rationales}   & Diverse & {Goal-driven}   & Interaction & {Counterfactual} &  \\
         & Answering  &  & Scenarios & Questions & of soft objects & Property Dynamics  \\
         \midrule
         IntPhys~\citep{riochet2018intphys}  & $\times$ &  $\times$ &  $\times$ & $\times$ & $\times$ & $\times$   \\
         ESPRIT~\citep{rajani2020esprit} & $\times$ &  $\times$ &  $\times$ & $\times$ & $\times$ & $\times$   \\
         Cater~\citep{girdhar2019cater} & $\times$ &  $\times$ &  $\times$ & $\times$ & $\times$ & $\times$  \\
         CoPhy\citep{Baradel2020CoPhy} & $\times$ &  $\times$ &  $\times$ & $\times$ & $\times$ & $\surd$  \\
         CRAFT~\citep{ates2020craft} & $\surd$ &  $\surd$ &  $\times$ & $\times$ & $\times$ & $\times$  \\
         CLEVRER~\citep{yi2019clevrer} & $\surd$ &  $\surd$ &  $\times$ & $\times$ & $\times$ & $\times$    \\
         Physion~\citep{bear1physion} & $\times$ &  $\times$ &  $\surd$ & $\times$ & $\surd$ & $\times$   \\
         ComPhy~\cite{chen2022comphy} & $\surd$ &  $\surd$ &  $\times$ & $\times$ & $\surd$ & $\surd$    \\ 
         ACQUIRED~\cite{acquired} & $\surd$ &  $\times$ &  $\surd$ & $\times$ & $\times$ & $\surd$    \\
         CRIPP-VQA~\cite{patel2022cripp}  & $\surd$ &  $\surd$ &  $\times$ & $\surd$ & $\times$ & $\times$  \\
        \midrule 
        \dataset (Ours) & $\surd$ &  $\surd$ &  $\surd$ & $\surd$ & $\surd$ & $\surd$  \\
        \bottomrule 
	\end{tabular}
	}	
% \vskip -0.1in
\end{table*}

\vspace{-0.5em}
\section{Related Work}
\vspace{-0.5em}
%\paragraph{Physical Reasoning.}
\paragraph{{Physical Reasoning.}}
Our work is closely related to Physical Reasoning benchmarks~\citep{rajani2020esprit,girdhar2019cater,Baradel2020CoPhy,bear1physion,li2022learning,li2022on}. We summarize the key features of these various benchmarks and compare them against our benchmark in table~\ref{tab:dataset_comparison}. Early benchmarks~\citep{riochet2018intphys,rajani2020esprit} simulate physical scenes with visual primitives and test models' physical intuition. Later, CLEVER~\citep{yi2019clevrer}, ComPhy~\citep{chen2022comphy}, and CRIPP-VQA~\citep{patel2022cripp} extend the simple visual primitives with natural language and asked questions about rigid bodies' collisions. Recently, Physion~\citep{bear1physion,tung2023physion++} provides more complex visual scenes and requires models to predict whether two objects will come into contact in future frames. As summarized in table~\ref{tab:dataset_comparison}, the proposed \dataset is the only benchmark that contains soft objects with different physical parameters and asks diverse language-based questions about dynamics in counterfactual and goal-planning scenarios.

\paragraph{{Video Question Answering.}}
Our paper is also related to Visual Question Answering (VQA) ~\citep{lei2018tvqa,zadeh2019social,jang2017tgif,chen2021grounding,wu2021star,ding2021dynamic,hong2023threedclr,chen2024visual,chen2023genome}, which mainly requires machine models to answer questions about a given image or video's content like visual attributes, actions, activity, and social events.
However, existing VQA datasets~\cite{zadeh2019social,xu2016msr,wang2019vatex,acquired,wang2024sok} still typically assess abilities in visual perception, recognizing objects, shapes, and colors, and understanding human-centric actions. In this paper, we aim to build a benchmark that evaluates AI models' comprehensive physical reasoning abilities.

\paragraph{{Physical Benchmarks for Soft Bodies.}}
Recently, there has been growing interest in the properties and dynamics of soft-bodied objects~\citep{xiang2020sapien,gan2020threedworld,macklin2014unified,xian2023fluidlab,haas2014history}. Much of the research has concentrated on creating simulations of deformable objects and fluids to advance robotic manipulation and cognitive experimentation. Leveraging simulation tools, we can simulate deformable objects and fluids with varying physical parameters, enabling collaboration with natural language for physical commonsense reasoning.

\section{Dataset}
The proposed \dataset dataset aims to assess the reasoning abilities of AI models across a wide spectrum of physical scenes of the continuum encompassing rigid bodies, soft bodies, and fluids, with massive physical properties treated as variables. In this section, we outline the dataset construction process. In Section~\ref{subsec:sim}, we describe the dataset's diversity in physical scenes and provide an introduction to each scenario. Section~\ref{subsec:question} explains the well-structured nature of our question dataset, encompassing properties and dynamics. In Section~\ref{subsec:stat}, we elucidate how a physical engine is employed to simulate diverse scenes with varying properties, the development of a question engine for question generation, and steps taken to mitigate dataset bias through statistical analysis.

\subsection{Diverse Physical Scenarios}
\label{subsec:sim}
We design four physical scenarios to visually illustrate various continuum physical dynamics and to study different physical behaviors across different object materials with varying physical properties.

\paragraph{{Diversity.}} The diversity of our video data arises from a wide range of materials with varying properties and changing dynamic phenomena. We include these materials and their physical properties in our scenarios compositionally, requiring models to obtain a deeper understanding of these scenes. First, this encompasses rigid bodies, ropes, cloths, plastoelastic balls, fluids, and articulated bodies, as well as their couplings, expanding our dataset diversity. For instance, Figure~\ref{fig:teaser} (c) portrays the coupling of deformable ropes and articulated pulleys. Second, a key feature that distinguishes the proposed \dataset dataset from existing benchmarks like CLEVRER~\cite{yi2019clevrer} and Physion~\cite{bear1physion}, is the inclusion of varying physical properties and commonsense concepts that are generally acknowledged by common people. These properties could scarcely be inferred through static images, involving mass, density, tension, friction, stretchiness, bending stiffness, elasticity, and plasticity. Such variation can alter the dynamics and generate differing future states, enriching the dataset with a wider range of observable physical behaviors. As an example, the liquid's position in Figure~\ref{fig:teaser}~(a) is jointly determined by its density and its interaction with the nearby liquids.

\paragraph{{Scenarios.}} We design four physical dynamic scenarios to comprehensively benchmark models' cognitive ability on the continuum, as shown in Figure~\ref{fig:dataset}. \\
\textbf{a) Liquid Dynamics.} In Figure~\ref{fig:dataset} A, we present a liquid hourglass-like device. Different liquids, each with unique colors and densities, are released from upper emitters, flowing through fixed sticks, altering directions, and finally reaching containers at the bottom. This setup reveals distinct behaviors arising from interactions between fluids with varied densities and dynamic trajectories. Our research focuses on investigating the physical properties and trajectories of these liquids. \\
\textbf{b) Cloths Manipulation.} As shown in Figure~\ref{fig:dataset} B, two cloth pieces with distinct stretching, bending, and frictional characteristics are pulled over objects, inducing potential collision events. The released fabric obstructs object views but outlines their shapes through deformations. Objects may topple if they surpass a height threshold or have low mass. This test assesses models' ability to discern fabric properties and predict spatial behaviors of concealed objects based on the dynamic 3D surface geometry of the fabric.\\
\textbf{c) Rope Pulley System.} In Figure~\ref{fig:dataset} C, a wall-mounted arrangement of pulleys, both movable and fixed and anchor points are depicted. Objects with varying masses interact, resulting in diverse motion patterns. The model's main goal is to identify tension distributions within this basic rope system. It must also recognize correlations or constraints among moving objects, such as coordinated loads and pulley rotations on a single rope. Additionally, the model is expected to infer numerical relationships between loads' masses and rope segment tensions.\\
\textbf{d) Soft Ball Dynamics.} Figure~\ref{fig:dataset} D illustrates a playground with colored obstacles and randomly placed pits. Plastoelastic balls with different deformation resistance and yield stress are launched from varying positions, undergoing dynamic movements, including bouncing and permanent deformation. Some balls may collide with obstacles and fall into pits. This experiment assesses the model's ability to accurately discern the elasticity and plasticity properties of soft bodies and predict their dynamic behavior.

\subsection{Diverse Structured Questions}
\label{subsec:question}

We categorize questions into two major groups: Physical Property Questions and Dynamics Questions. Figure~\ref{fig:dataset} shows the question types of the four scenarios. Sample templates are provided in Table~\ref{tab:fluid_q}, ~\ref{tab:rope_q}, ~\ref{tab:cloth_q}, and ~\ref{tab:ball_q} in the Appendix.

\paragraph{{Physical Property Questions.}}
We formulated a set of physical property questions across four distinct scenarios. We pose questions about our chosen physical properties which can only be answered by observing object dynamics and interactions, neither static images nor single object behaviors. These questions can be answered with a brief phrase. Models are expected to deduce physical properties based on input video data, which requires physical common sense. Besides, we also inquire about the visible physical properties of objects, such as colors, shapes, and existences, which are shown in static frames. 

\paragraph{{Dynamics Questions.}}
Dynamic questions can be further categorized into three types: counterfactual, goal-driven, and predictive, respectively concerning potential outcomes of changed conditions, strategies for specific objectives, and predictions about the future. In the fluid and ball scenarios, we crafted questions covering all three types, anticipating models to develop a comprehensive understanding of these scenarios through diverse question templates. For rope and cloth scenarios, we selectively assess a subset of dynamic question types due to scenario complexity. In the rope scenario, only counterfactual and goal-driven questions are included. In the cloth scenario, exclusively predictive questions prompt the model to anticipate outcomes not directly visible under the cloth cover.
To increase cognitive challenge, we've designed multiple-choice questions with more than two but fewer than five answer choices, requiring models to provide a binary prediction for each option.

\begin{figure*}[htp]
\begin{center}
\centerline{\includegraphics[width=1\linewidth]{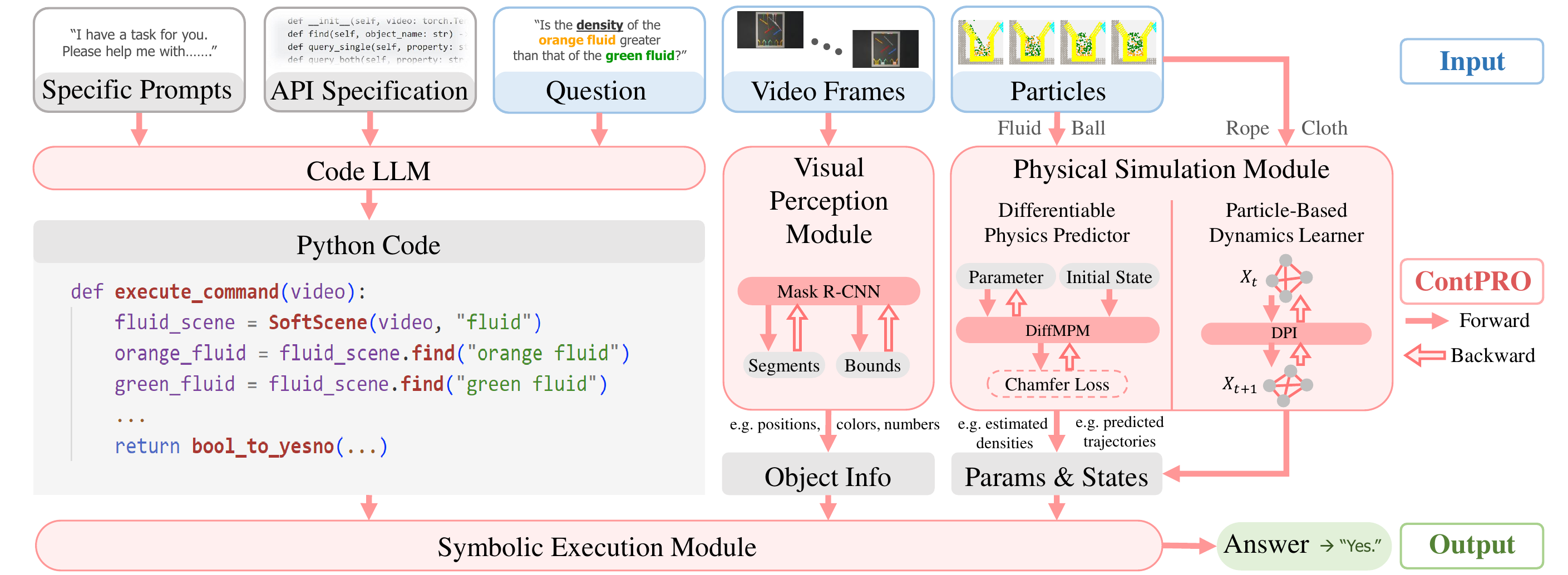}}
\vskip 0.1in
\caption{
{
The architecture of the \textbf{\model} model.
With questions, predefined APIs, and specific prompts, an LLM will play as a program parser that translates questions into code snippets. The visual perception module predicts objects' location and static attributes. The physical simulation module predicts dynamics. The symbolic execution module executes the code snippet to output the answer.} 
}
\label{fig:sopro}
\end{center}
\end{figure*}

\subsection{Generation Setup and Statistics}
\label{subsec:stat}
\paragraph{{Video Generation.}}  
We used the Unity engine~\citep{haas2014history}, an efficient platform, to simulate and render videos. We follow a bottom-up approach to generate videos and their annotation, involving the following sequential steps:\\
\textbf{a) Sampling.} Randomly select scene layouts, camera parameters, and initial conditions to create a diverse set of scenarios.\\
\textbf{b) Initialization.} Place and configure all objects within the scenarios.\\
\textbf{c) Pre-simulation.} Conduct a preliminary simulation to evaluate whether the obtained simulation results align with the expected data distribution.\\
\textbf{d) Rendering.} Generate high-quality videos with configured cameras.\\
\textbf{e) Post-simulation.} Carry out multiple simulations under varying conditions and record the simulation outputs. \\
\textbf{f) Output.} Produce rich sensor data and annotation information, encompassing original video, segmentation, bounding boxes, particles, meshes, collision events, configurations, and other simulation raw data required for question generation. We will provide more details in Section~\ref{app:vid} in the Appendix.

\paragraph{{Question Generation.}}
We develop a question engine to generate question-answering pairs step by step:\\
\textbf{a) Template Design.} Create massive question and option templates. \\
\textbf{b) Sampling.} Retrieve the simulation results, combine the properties of possible objects with predefined templates, sample questions, and options accordingly, and determine correct answers. Target objects possess unique names described by visual attributes like color, shape, orientation, and mobility. \\
\textbf{c) Re-Sampling.} Ensure a balanced distribution of answers among the options to prevent answer bias. 

\paragraph{{Question Statistics.}} The video content prompted the formulation of numerous questions, with each video featuring one property question and two dynamics questions, except for the rope scenario, which included two property-related questions and two dynamics questions. A total of 2,000 questions were generated for the rope scenario, and 1,500 questions were created for other scenarios. The dataset encompasses 6,500 questions derived from 2,000 videos. We divided the dataset into three subsets: 50\% for training, 20\% for validation, and 30\% for testing. Across the entire dataset, 20\% of questions are counterfactual, 11\% are goal-driven, 22\% are predictive, and the remaining 46\% are related to various physical property questions. Further details on the distribution of each question type and templates for each scenario can be found in Section~\ref{app:tqmp} in the Appendix.

\section{\model}\label{app:model}

{
Inspired by prior work~\citep{yi2018neural,yi2019clevrer}, we propose an oracle neural-symbolic framework named \model for \dataset. 
As shown in Figure~\ref{fig:sopro}, we decompose the question-answering task into four main modules, video perception, physical simulation, program parser, and symbolic execution.
Given a raw video, the video perception module detects the objects and their associated static attributes with MASK-RCNN detector~\cite{he2017mask}. The physical simulator takes point clouds as input and predicts objects' dynamics in different scenarios with dynamic prediction models~\cite{jiang2015affine,li2018dpi}. The program parser translates the question query into executable programs with a large language model (LLM). Based on the object attributes and dynamics, the symbolic executor executes the programs to get the answer to the question.
}

{
Compared with previous models, \model is the first model that marries LLMs with particle-based physical dynamic models, which eliminates the need for handcraft design and in-domain training of the program parser and enjoys precise dynamic prediction.
As it requires point clouds as input, we call it an oracle model. 
} 

{
\paragraph{{Video Perception.}} 
The video perception module is supported by a MASK R-CNN~\citep{he2017mask} to densely detect objects' location in each frame and associated static attributes like color and material. We take the ResNet-50~\cite{he2016deep} as the backbone and fine-tune the network with data from the training set of all four scenarios until converges. 
}

{
\paragraph{{Physical Simulation.}}
We choose DPI-Net~\citep{li2018dpi} for dynamic prediction for rope and cloth as it has shown reasonable dynamic prediction abilities in different materials~\cite{chen2022comphy,bear1physion}. We also observe that the Material Point Method (MPM)~\cite{sulsky1995application} can estimate physical properties and dynamics better for fluids and objects of varying plasticity with its differentiable mechanism during inference. Thus, we adopt MPM to predict dynamics for scenes of fluid and soft balls.
}

{For DPI-Net, we train it with the cloth and rope data from the training set. For the MPM model, we first initialize the physical properties of the object with a fixed value and gradually optimize its value with gradient descent. We set the optimized loss target to be the chamfer loss between the groups of 3D points and the predicted points.}

{
\paragraph{{Large Language Model as Program Parser.}} 
Traditional neuro-symbolic models~\cite{andreas2016neural,yi2018neural} usually train a domain-specific sequence-to-sequence model to translate the natural language query into executable programs, which requires manual implementation of each symbolic operator and show problems in generalizing to questions out of the training set distribution. Motivated by the recent ViperGPT~\citep{surismenon2023vipergpt}, we utilize the large language model, ChatGPT~\citep{ouyang2022training} as the language parser. We further develop a set of dynamics modules and visual perception modules serving as APIs for solution generation. With the provided API access and a pre-defined physical reasoning prompt, we leverage ChatGPT to generate Python code that can be directly executed and interpreted. This module bridges the gap between language comprehension and physical concept understanding.
}

\paragraph{{Program Execution.}}
After extracting objects' static attributes, physical properties, and dynamic trajectories and parsing the natural language query into an executable program, we execute the program with object states as input and output the predicted answer. We provide more model details in Section~\ref{sec:ContproDetails} in the Appendix.

\section{Experiments}
\newcommand{\Frst}[1]{\textcolor{red}{\textbf{#1}}}
\newcommand{\Scnd}[1]{\textcolor{blue}{\textbf{#1}}}
\newcommand{\Thrd}[1]{\textcolor{orange}{\textbf{#1}}}
\begin{table*}[t]
    \caption{Physical reasoning on \dataset. We list all question families, \textbf{P}roperty, \textbf{C}ounterfactual, \textbf{G}oal-driven and \textbf{P}redictive questions. Accuracy is reported with per \textbf{Opt}ion and per \textbf{Ques}tion. \Frst{Red} text, \Scnd{blue} and \Thrd{orange} text indicates the first, second, and third best result.}
 \vspace{0.1in}
    \resizebox{1\linewidth}{!}{
    \centering
    \setlength{\tabcolsep}{2pt}
    \renewcommand{\arraystretch}{1.25}
	\begin{tabular}{l|l|ccc|ccccc|cc|cc|c|c}
            
            \shline
            \multirow{2}{*}{Subset} & \multirow{2}{*}{Settings} & \multicolumn{3}{c|}{Blind Models} & \multicolumn{5}{c|}{Visual Models} & \multicolumn{2}{c|}{Physical Models} & \multicolumn{2}{c|}{MLLMs} & \multirow{2}{*}{\model} & \multirow{2}{*}{Human} \\
            
           & & RND & FRQ & B-LSTM & C-LSTM & MAC & HCRN & ALPRO & Violet & PhyDNet & PIP & Gemini & GPT-4V &  &  \\
            \shline
            \multirow{5}{*}{Rope} & Prop. & 30.0 & 53.3 & 54.7 & 52.7 & 53.3 & 51.7 & \Thrd{60.7} & 51.7 &  59.0 & 31.5 & 35.5 & 48.0 & \Scnd{71.4} & \Frst{84.7} \\
             & C-Opt.                     & 51.3 & 51.6 & 74.0 & 74.0 & 74.2 & 74.3 & \Thrd{76.2} & 76.0 & \Scnd{77.7} & 75.2 & 48.2 & 42.0 & 75.6 & \Frst{90.2} \\
             & C-Ques.                    & 14.7 & 19.0 & 46.0 & 45.0 & 39.8 & {48.1} & \Scnd{50.7} & 43.1 & 47.9 & 48.3 & 12.0 & 11.3 & \Thrd{48.8} & \Frst{75.0}  \\
             & G-Opt.                     & 55.2 & 49.7 & 47.4 & 51.2 & 50.3 & \Thrd{56.0} & 46.2 & 55.2 & 54.4 & 50.6 & 51.6 & \Scnd{57.0} & 55.8 & \Frst{91.9} \\
             & G-Ques.                    & 4.5 & \Thrd{11.2} & 7.9 & 6.7 & 6.7 & 2.3 & 1.1 & 1.1  & 5.6  & 2.2 & 10.3 & \Scnd{12.1} & 2.3 & \Frst{84.0}  \\
            \shline
            \multirow{7}{*}{Fluid} &Prop. & 33.3 & 52.7 & 49.3 & \Thrd{54.0} & 30.0 & 52.7 & 48.0 & 50.9 & 51.3 &  37.0 & 10.0 & 25.0 & \Frst{78.0} & \Scnd{75.8} \\
             & C-Opt.                     & 52.9 & 57.9 & 56.1 & 55.0 & 56.5 & 52.6 & 56.8 & \Thrd{60.4} & 59.5 & 49.1 & 47.3 & 53.3 & \Scnd{75.7} & \Frst{82.5} \\
             & C-Ques.                    & 6.0  & \Thrd{17.2} & 7.8  & 8.6  & 6.9  & 4.3  & 6.0  & 1.7  & 10.3 & 6.0 & 5.1 & 5.1 & \Scnd{36.2} & \Frst{60.6} \\
             & G-Opt.                     & 59.9 & 63.1 & 57.3 & 57.3 & 51.2 & \Thrd{67.7} & 62.7 & 67.3 & 55.9 & \Thrd{67.7} & 44.4 & 53.8 & \Frst{77.3} & \Scnd{75.0} \\
             & G-Ques.                    & 7.5  & 36.3 & 22.5 & 22.5 & 17.5 & \Thrd{41.3} & 32.5 & 41.2 & 40.0 & \Thrd{41.3} & 11.3 & 7.5 & \Scnd{60.0} & \Frst{64.3} \\
             & P-Opt.                     & \Thrd{53.8} & 50.1 & 51.4 & 51.4 & 53.5 & 50.6 & \Thrd{53.8} & 53.2 & 51.7 & 45.5 & 52.4 & 50.0 & \Frst{90.1} & \Scnd{73.9} \\
             & P-Ques.                    & 4.8  & 12.5 & 12.5 & 12.5 & 12.5 & 1.9 & 12.7 & 3.8  & 4.8  & 3.8 & 5.8 & \Thrd{13.0} & \Frst{68.3} & \Scnd{42.9} \\
            \shline
            \multirow{3}{*}{Cloth}       &Prop. & 46.7 & 41.3 & 56.7 & 46.7 & \Thrd{59.3} & 52.0 & 48.0 & 55.0 & 58.7 & 54.0 & 42.0 & 49.0 & \Scnd{60.0} & \Frst{81.4} \\
             & P-Opt.                           & 52.2 & 61.7 & 55.2 & 67.5 & 57.9 & 62.0 & \Scnd{68.8} & \Thrd{68.2} & 63.5 & 61.6 & 50.1 & 53.0 & {64.7} & \Frst{79.6} \\
             & P-Ques.                          & 46.0 & 56.7 & 42.3 & \Thrd{57.3} & 50.7 & 56.3 & \Thrd{57.3} & 55.7 & 47.3 & 46.3 & 43.0 & 47.5 & \Scnd{60.0} & \Frst{77.3} \\
            \shline
            \multirow{7}{*}{Ball} & Prop.       & 53.5 & 52.0 & 45.3 & \Scnd{54.7} & 48.0 & 43.3 & 48.0 & 48.0 & 52.7 & \Thrd{54.0} & \Thrd{54.0} & 45.0 & \Thrd{54.0} & \Frst{76.9} \\
             & C-Opt.                           & 53.6 & 65.8 & 66.7 & 64.2 & 66.1 & 65.3 & 63.9 & 65.6 & \Thrd{67.2} & 63.7 & 60.9 & 66.7 & \Scnd{71.6} & \Frst{93.9} \\
             & C-Ques.                          & 30.4 & \Thrd{48.7} & 43.4 & 41.8 & 3.3  & 28.7 & 40.2 & 41.8 & 44.3 & 24.6 & 29.6 & 46.9 & \Scnd{57.4} & \Frst{90.9}  \\
             & G-Opt.                           & 55.9 & 52.1 & 53.3 & 54.1 & \Thrd{58.1} & 57.0 & 56.3 & 57.4 & 57.4 & 54.1 & 54.1 & 51.4 & \Scnd{68.1} & \Frst{89.7} \\
             & G-Ques.                          & 30.2 & 38.5 & 16.7 & 20.0 & 18.9 & \Thrd{38.9} & 4.4 & 21.1 & 21.1 & 22.2 & 24.6 & 18.0 & \Scnd{52.2} & \Frst{84.6} \\
             & P-Opt.                           & 50.6 & 67.8 & \Thrd{68.9} & 67.4 & 64.4 & 61.7 & 65.2 & 64.4 & 67.4 & 62.9 & 51.7 & 45.4 & \Frst{92.4} & \Scnd{72.5}\\
             & P-Ques.                          & 25.9 & \Thrd{51.7} & 45.5 & 45.5 & 46.6 & 1.1  & 3.4 & 2.3  & 17.0 & 6.8 & 25.9 & 17.2 & \Frst{88.6} & \Scnd{58.8}\\
            \shline
	\end{tabular}
	}
	\label{tab:results}
\end{table*}

In this section, we evaluate the ~\dataset dataset. We first introduce the experimental setup and then analyze the performance of different models.

\subsection{Experimental Setup}
For simplicity, each physical property question is regarded as a classification task among all possible answers. Each dynamic question is treated as a binary classification task for each question-choice pair. For dynamic questions, we report the accuracy of per option per question. A question is correct only if all choices in this multiple-choice question are correctly answered.

\paragraph{{Blind Models.}} This family of models includes baselines that only rely on question input, to analyze language biases in ~\dataset. 
\textbf{RND} chooses at random a possible answer, or randomly selects between true-false binary answer pairs for every multiple-choice question.
\textbf{FRQ} selects the most frequent answer based on the question type.
\textbf{B-LSTM} utilizes an LSTM~\citep{hochreiter1997long} to encode the questions only and predict answers.

%\paragraph{Visual Models.}
\paragraph{{Visual Models.}}
These models incorporate both visual and language representations for answering questions. 
\textbf{C-LSTM} extracts video features via ResNet-50 convolutional neural network (CNN)~\citep{he2016deep} on 25 sampled frames of videos and averages them over time as the visual input. We concatenate this visual input with the question embedding from the last hidden state of LSTM to predict answers. 
\textbf{HCRN}~\citep{le2020hierarchical} uses conditional relational networks to learn relations hierarchically in the video, as well as the questions. 
\textbf{MAC}~\citep{hudson2018compositional} has competitive results on previous datasets, which uses a co-attention mechanism to model both textual and visual information.  
\textbf{ALPRO}~\citep{li2022align} is a popular model pre-trained on video-text corpus and achieved state-of-the-art results on several video-language datasets.
We fine-tune ALPRO on our dataset based on the official pre-trained checkpoint.

\paragraph{{Physical Models.}} These specialized models are trained for physical reasoning. 
\textbf{PhyDNet}~\cite{leguen20phydnet} is a two-branch deep architecture, which explicitly disentangles PDE dynamics from unknown complementary information.
\textbf{PIP}~\cite{duan2021pip} utilizes a deep generative model to model mental simulations, in order to predict future physical interactions.
To evaluate our benchmark on these physical baselines, we first generated object masks based on each question and fed them into models, together with the video features. For the open-ended questions, we added a fully connected layer to predict the answer labels with a cross-entropy loss. 

{\paragraph{Multimodal Large Language Models (MLLMs).} This model family represents the cutting edge in solving vision-language problems. We consider two pioneering models, \textbf{GPT-4V(ision)}~\citep{openai2023gpt4} and \textbf{Gemini}~\citep{team2023gemini}, which have demonstrated extraordinary performance.

\subsection{Evaluation of Physical Reasoning}
We summarize the performance of all baselines in Table~\ref{tab:results}. The results show that different models exhibit distinct performance variances across different question types and scenarios. This indicates that ~\dataset can evaluate models' physical reasoning capabilities in different dimensions.

\paragraph{{Performance of Blind Models.}} {Blind models operate and respond solely to textual data, reflecting the quality and structure of question design. Generally, these models have weaker performance than other families of models, showing the importance of cooperating visual information to handle the questions in \dataset. We also observe that blind models perform similarly to other model families in some scenarios like the goal-driven questions of the rope scenario. We think the reasons are that these dynamic questions are too challenging for current machine models, which require the understanding of physical commonsense and predict information that is not directly observable. Note that human beings can still easily achieve high performance.} 

\paragraph{{Performance of Visual Models.}} {Visual models, which integrate both visual and language representations, exhibit relatively consistent performance across various questions.
Among all dynamic questions, they excel on the rope's counterfactual and cloth's predictive, but fall short on goal-driven questions. The inherent complexity of goal-driven questions, which require reverse reasoning based on the goal, may account for its bad performance.
Among different scenarios and visual models, ALPRO distinguishes itself by its robust overall performance, notably in cloth and rope, which shows the advantages of large-scale video-text pre-training and alignment, emphasizing its effectiveness in complex visual reasoning.
Despite these advancements, no visual model has yet achieved top accuracy in all scenarios, underscoring the challenge and significance of our \dataset.
} 

\paragraph{{Performance of Physical Models.}}
Physical models are specialized models designed for special physical tasks. These models achieve competitive overall performance and excel in some settings, such as PhyDNet on rope's and ball's counterfactual, and PIP on fluid's goal-driven. However, these specialized models also have limitations in some scenarios, such as cloth. We hypothesize the reasons are that these models are mainly designed for physical reasoning tasks with simple visual primitives, like sphere collision and movement. However, our dataset focuses on continuum objects in diverse environments and different question types, which makes it difficult for these models to grasp the physical rules behind the scenarios. 

\paragraph{{Performance of Multimodal Large Language Models.}}
Compared with other baselines, both foundation models, Gemini and GPT4-V, fall short in cloth and fluid questions in both property and dynamics levels, showing that they fail to perceive highly deformable objects. However, they perform the best in certain question settings of rope and ball scenarios. Note that these models have never been trained on \dataset and their visibility is limited to discrete frames in our setup. {We think the reason is that objects in rope and ball, \eg cubes and spheres, are more common to foundation models than those in fluid and cloth, \eg different colored liquids and clothes.} {For example, GPT-4V shows its capabilities in counting objects and perceiving object colors, which probably raises the rope property evaluation score.} {MLLMs also distinguish themselves by their complex reasoning capabilities, accounting for high accuracy in rope's goal-driven questions.} We argue that current foundation models lack the necessary capability to infer physical properties and predict the dynamics of complex scenes with soft objects and fluid.

\paragraph{{Performance of \model.}} {\model excels significantly over other machine models on most questions, particularly on property inference and predictive questions. 
Notably, its accuracy on predictive questions even surpasses human performance. A possible explanation is that we adopt MPM for fluid and ball scenarios, which can precisely predict objects' positions in the short term, while humans tend to give vague estimations. Such an approach also accounts for its superior performance on the fluid's goal-driven questions. 
In addition, \model demonstrates exceptional performance on fluid property inference, achieving an accuracy of 96.9\% on ``stick number" and 63.5\% on ``density". We believe the reason is that we have utilized a fine-tuned Mask R-CNN predictor to identify sticks in videos. 
For rope and ball scenarios, we employ DPI-Net, a GNN-based simulator, which cannot exhibit absolute advantages over visual models.
} 

\paragraph{{Human Performance.}} 
We randomly sampled some video-question pairs from the test set to assess the human ability to comprehend the physical properties and dynamic events presented in both video and textual descriptions. 
To evaluate human performance on ~\dataset, 16 people participated in the study. Participants were required to have fundamental English reading skills and a basic physical knowledge background.
First, each participant was asked to select a scenario randomly, after which they were presented with distinct video-question pairs. Participants were instructed to answer with a phrase when presented with physical property questions, while for dynamics questions they were required to provide a binary true-false response from available choices. We obtained 460 valid human answers encompassing all scenarios and question types within ~\dataset. We can observe from Table~\ref{tab:results} that it beats visual models and foundation models in all scenarios. This shows the fundamental ability and strength of humans to perform visual reasoning and inference from videos.

\paragraph{{Evaluation Conclusion.}}
The strong human results demonstrate that humans maintain a strong capacity to comprehend both videos and questions, make physical property inferences from given videos, and predict and reason counterfactual hypotheses concerning unseen information. Machine model results show that even state-of-the-art models struggle with answering these physical questions. 
This indicates that our dataset poses a significant challenge for vision-language models to achieve similar basic physical video understanding ability with human beings. {We also propose an oracle model to demonstrate the potential to combine recent large language models with traditional particle-based dynamic simulation for effective physical reasoning.}

\section{Limitations}

Our proposed benchmark, \dataset, aims to complement existing physical reasoning benchmarks by encompassing diverse physical property inference across various scenarios and predicting corresponding dynamics. However, \dataset still has limitations.

\noindent\textbf{Language Diversity.} While the synthesized questions generated by the question engine can effectively test AI models' physical reasoning capabilities across diverse scenarios involving different objects, the language diversity remains limited. The current set of questions relies on a predefined vocabulary, resulting in a gap compared to natural language.

\noindent\textbf{Scenario Complexity.} We have carefully designed four distinct scenarios featuring various objects (\eg, solids, ropes, clothes, and fluids). However, real-world physical interactions can be considerably more complex, involving additional objects and physical factors not currently included in the dataset.

\section{Conclusion}
We introduced the \datasetFull~(\dataset), a pioneering benchmark for assessing machine models in physical reasoning of the continuum, especially for soft bodies and fluids. This benchmark broadens the scope by covering various physical property inferences for soft bodies across dynamic contexts and predicting their dynamics. Our dataset has enabled the development of AI models with human-like reasoning abilities, comprehending both visual attributes and complex physical properties of objects while solving problems.
Despite progress, our evaluation of AI models revealed an ongoing challenge: they struggle to perform well on our benchmark, highlighting their limited physical commonsense for the continuum, especially soft bodies, and fluids. 
We foresee the \dataset driving progress in AI perception and reasoning, bridging the gap between human and machine intelligence in the physical world.
%\newpage
\section*{Acknowledgement}
This work was supported by DSO grant DSOCO21072. We would also like to thank the computation support from AiMOS, a server cluster for the IBM Research AI Hardware Center.

\section*{Impact Statement}
This paper presents work whose goal is to advance the field of Physical Commonsense Reasoning. We believe our work is useful for 1) evaluating the performance of current existing machine learning models for physical reasoning, and 2) facilitating researchers to develop more powerful AI models with physical commonsense. There are no potential negative societal consequences of our work that we feel must be specifically highlighted here.

% % In the unusual situation where you want a paper to appear in the
% % references without citing it in the main text, use \nocite

\bibliography{example_paper}
\bibliographystyle{icml2024}

\clearpage
\section{Dataset Details}

\subsection{Video Details.}\label{app:vid}

For each simulation trial, we produce two primary sets of data: sensor output and semantic annotation. The sensor output provides a comprehensive 4D state description of objects at various levels. In contrast, the semantic annotation contains pre-processed data designed to facilitate the question-generation phase. 

\subsubsection{Sensor Data Structure.}

Within the simulation pipeline, we produce sensor data across multiple modalities listed in Figure~\ref{fig:sensoranno}, including RGB-rendered images in Full HD ($1920\times1080$) resolution, object-level data (encompassing bounding boxes, segmentations, positions, rotations, and scales), point-level data (comprising meshes and particles), and event-level data (detailing collision or touch events). The generated meshes illustrate the sampled surface shapes of both rigid and soft objects, prepared for subsequent voxelization. Unlike the other two scenarios, the fluid and rope scenarios necessitate the re-sampling of meshes in every individual frame. This results in temporal independence for the vertices. Yet, within this context, particle outputs signify tracked points on the objects, preserving correlations between successive frames. Given that voxel data (which is temporally invariant) is derived from the voxelization of meshes, the dataset offers both temporally correlated and independent 4D data.

\begin{figure*}[ht]
\vskip 0.2in
\begin{center}
\centerline{\includegraphics[width=\linewidth]{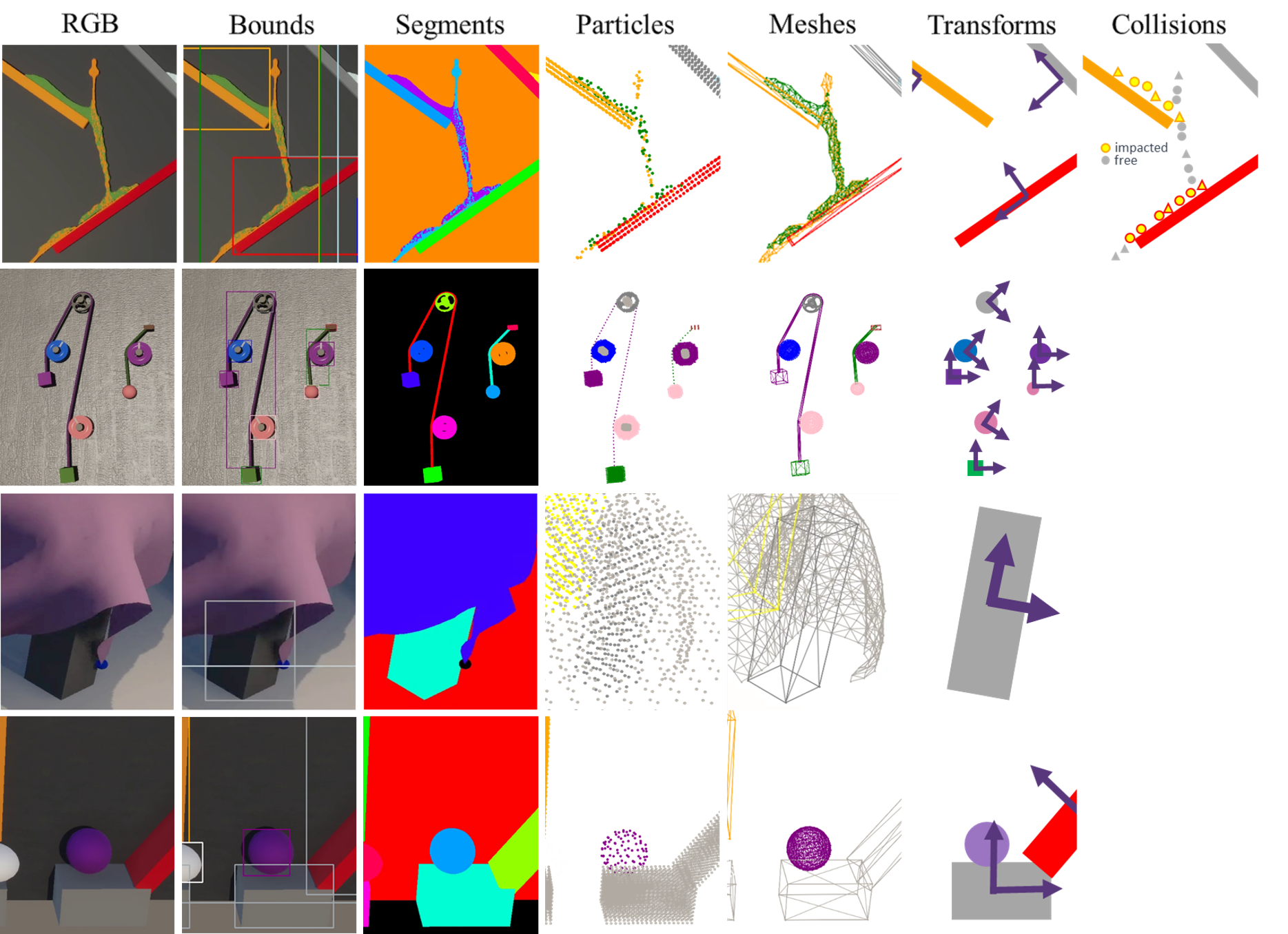}}
\caption{Sensor data outputs are multimodal, depicting the 4D states of objects across various levels, ranging from object-level, point-level to event-level.}
\label{fig:sensoranno}
\end{center}
\vskip -0.2in
\end{figure*}

\subsubsection{Annotation Data Structure.}
For each scenario, we produce comprehensive annotation data that includes camera extrinsics/intrinsics, sampled parameters, and properties of the sampled objects and layouts. Additionally, post-processed simulation data from both the pre-simulation and post-simulation stages are documented. To be specific:

\paragraph{Fluid.} Object details such as name, color, and transforms are stored. For fluid objects, properties like densities, viscosity, surface tension, and emitted positions are added. Particle statistics in each container, collision statistics on each stick, and collision paths for each particle are recorded for both pre-simulation and post-simulation stages, and are meticulously categorized by fluid types.

\paragraph{Rope.} Fundamental elements of each pulley group, such as pulley, rope, fixed endpoint, cube, and sphere, are outlined at both the individual rope and group levels. A group refers to a collection of objects with interdependent mechanics, like two sets of objects on ropes connected to a specific movable pulley. Initial properties such as mass, color, shape, mobility, pose, and subsequent simulation results like motion direction of movable objects and tension in rope segments are annotated.

\paragraph{Cloth.} Sampled cloth properties—stretching compliance, bending compliance, and friction level—are provided. Basic properties of each rigid object and their simulation results, which include object-cloth and object-object collision events, contact relationships, and tension values in the cloth's final frame, are stored.

\paragraph{Ball.} The framework documents sampled properties of all rigid bodies and soft balls. For plastoelastic balls, simulation results, including the pits they settle into, are captured.

\subsubsection{Physical Video Diversity}

In the video part of our dataset, we have generated a substantial volume of videos, physical parameters, and objects for diverse questions. To provide a more detailed breakdown, we categorize videos by scenario. Each scenario contains 500 videos of fixed lengths: 250 frames for fluid, 150 for rope, 145 for cloth, and 120 for ball. Given the diverse responses in the VQA generation phase, we employed randomization for several configuration parameters during the simulation initialization. Beyond general scene arrangements like camera, lighting, and backgrounds, unique configurations pertain to each scenario:

\paragraph{Fluid.} Fluid density factors into multi-fluid interactions. Striving for diverse results, the number of fluid emitters and containers, the positions, poses, scales of obstructive sticks, and object colors are randomized. Fluid densities, chosen from a preset pool, should ensure discernible stratification in fluid interactions.

\paragraph{Rope.} The rope-pulley system layout, rope link lists, and entanglement methods are pre-set to allow varied connections between adjacent objects. Filtering steps identify simulations that provide diverse and aesthetically pleasing configurations. Attributes such as color, shape, load mass, load movability for loads, ropes, fixed endpoints, and pulleys are randomized prior to simulation.

\paragraph{Cloth.} Parameters like stretching compliance, bending compliance, and friction rate are drawn from a predetermined pool, ensuring cloth dynamic differences discernible to humans. Other items, such as pillars and plates, undergo random scaling and positioning. Cloth movement speeds and paths vary, aiming for diverse collision outcomes. Rigid object masses are also randomized to diversify collision event predictability.

\paragraph{Ball.} Deformation resistance and plasticity yields are sourced from a set value range to highlight differing properties. Floating wall positions and poses are constrained to specific zones to intensify collision events in videos, leading to varied outcomes during and post-video.

\begin{figure*}[ht]
\begin{center}
\centerline{\includegraphics[width=0.86\linewidth]{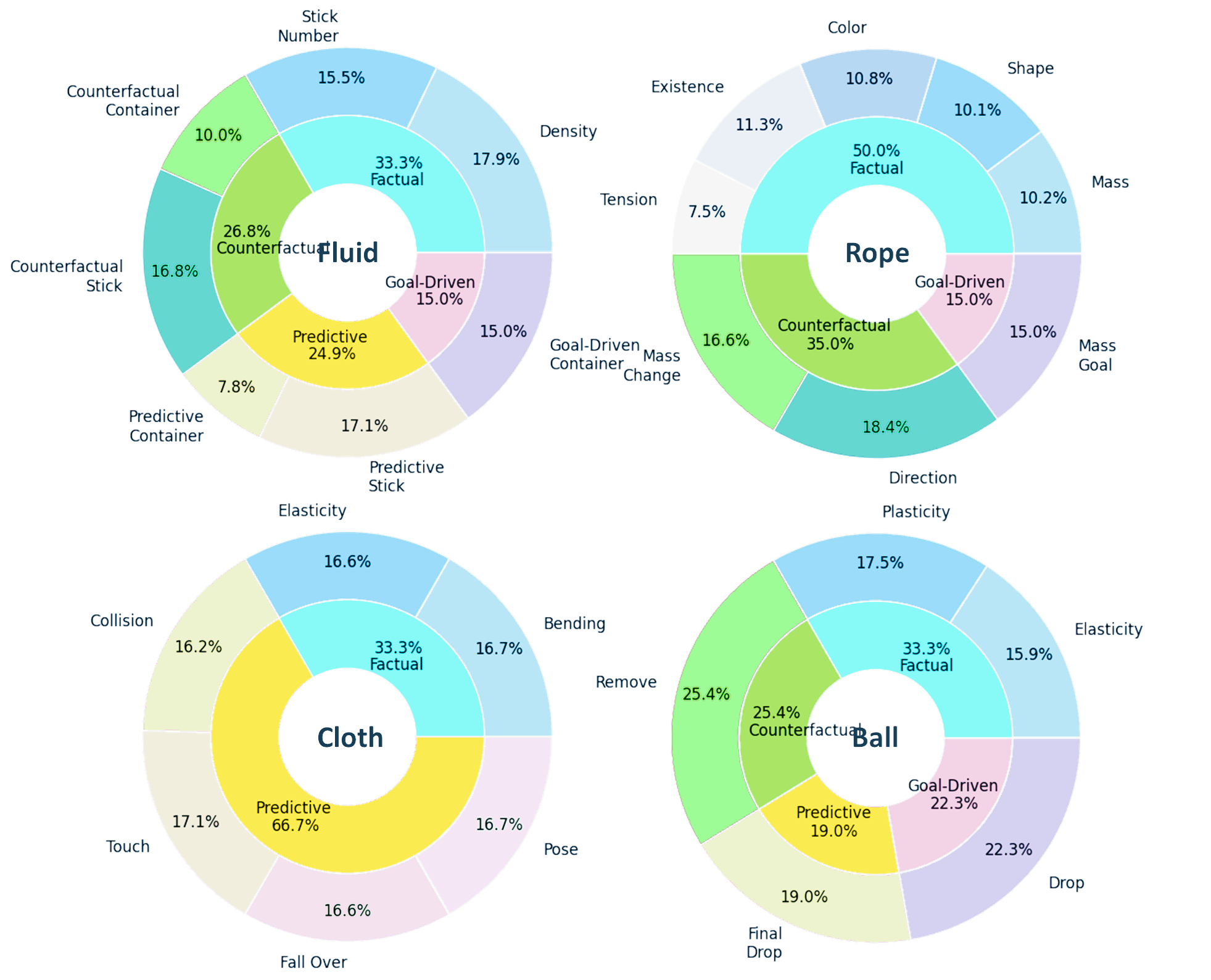}}
\caption{
Question distribution of fluid, rope, cloth, and ball scenarios.
}
\label{fig:distri}
\end{center}
\end{figure*}

\subsection{Question Details.}\label{app:tqmp}

\subsubsection{Question Distribution}\label{app:qdis}

In this section, we visualize the distribution of different types within different scenarios, including rope, fluid, cloth, and ball in Figure~\ref{fig:distri}. We balance the number of each question type. We also provide a comparison of question type distribution with ComPhy and CLEVRER in Figure~\ref{fig:contphy_qtype_pie}.

\subsubsection{Question Templates and Examples}
We show all question templates and examples from four scenarios in Table~\ref{tab:fluid_q}, ~\ref{tab:rope_q}, ~\ref{tab:cloth_q}, and ~\ref{tab:ball_q}. All the symbols are defined in Table~\ref{tb:sym}. When generating questions using templates and symbols, we balance the distribution and frequency of each symbol and answer to avoid language bias.

\begin{table}[hbp]
\caption{{Detailed explanation of symbols that we use in question generation with question templates.}}
\label{tb:sym}
\vskip 0.15in
\centering
\begin{tabular}{cl}
\toprule
Symbol  & Explatations \\
\midrule
\multirow{2}{*}{\_CLR\_}      & 
\textcolor{blue}{blue}, 
\textcolor{black}{black}, 
 \textcolor{brown}{brown},
 \textcolor{cyan}{cyan},
\textcolor{gray}{gray},
 \textcolor{green}{green},
 \textcolor{pink}{pink}, 
 \\ 
 &  \textcolor{orange}{orange},
\textcolor{purple}{purple}, 
\textcolor{red}{red}, 
\textcolor{yellow}{yellow}, 
\textcolor[RGB]{173,216,230}{light blue},
  \textcolor[RGB]{207,207,196}{white}
  \\
    \midrule
\_SHP\_   & solid, hollow
 \\
  \midrule
\_OBJ\_   & plate, pillar, cube, sphere, pulley, rope
 \\
\midrule
\_CMP\_   & greater than, less than, harder, easier, equal to
 \\
       \midrule
\_FAC\_   & twice of, half of
 \\
  \midrule
\_POS\_   & left, right
 \\
   \midrule
\multirow{2}{*}{\_ENT\_}   & move up, move down, \\
&rotate clockwise, rotate anti-clockwise
 \\
\bottomrule
\end{tabular}
\end{table}

\subsubsection{Logical Steps to Infer Answers}

Similar to other synthesized questions in previous research~\cite{patel2022cripp, yi2019clevrer}, we can get the logical steps, \ie reasoning operators, that lead to the answer. To provide more information about the benchmark, we show the exemplar logical steps for examples of each question type and calculate their statistics in Table~\ref{tab:fluid_q}, ~\ref{tab:rope_q}, ~\ref{tab:cloth_q}, ~\ref{tab:ball_q}. We can see that most types of questions have two or three logical steps, which involve diverse capabilities for querying objects' visual attributes, physical properties, and dynamics based on the physical properties of solid objects, soft objects, and liquids.

\subsubsection{LLM-based Question Rephrasing}

To enhance the diversity of \dataset question data, we use the large language model Gemini as an automatic rephrasing tool, to help rephrase the question texts. The instruction prompt is listed in Table~\ref{tb:prompt_rephrase}. 

We provide more statistical data about the question dataset before and after rephrasing and the comparison between \dataset and two former works in Table~\ref{tab:apdx_stat}. We use TTR (Type-Token Ratio) and Word Distribution (See Figure~\ref{fig:contphy_wdist_pie}) to evaluate the \textbf{lexical diversity}. We report each sentence's average length and variance to evaluate the \textbf{syntactic diversity}. We also report the question type number, and detailed question type distribution (See Figure~\ref{fig:contphy_qtype_pie}) to evaluate the \textbf{question type diversity}. Also, the F-K (Flesch-Kincaid) Grade Level is considered a \textbf{readability score} for reference. 

We provide a comprehensive evaluation of different prompting methods. Please refer to Section~\ref{app:exp_prompt} for more details.

\subsubsection{Word Distribution}

We also visualize the distribution of the word in our questions within different scenarios in Figure~\ref{fig:contphy_wdist_pie}. For each scenario, we show the word distribution before and after LLM rephrasing. Results show that the questions are more diverse and the distribution is more balanced after rephrasing. We also compare with two previous works, including ComPhy and CLEVRER.

\section{The Continuum: Liquids, Soft Bodies, Rigid Bodies, and Articulated Bodies}\label{app:continuum}

In this section, we consider the physical concept of the continuum.
Previously, physical datasets mainly focused on simple visual primitives of rigid bodies, such as cubes and spheres.
In our \dataset, we extend this success to a broader concept, the continuum. The continuum encompasses various bodies such as liquids, soft materials (\eg, soft balls, cloth, and ropes), rigid bodies (\eg, cubes, pillars, plates and spheres), and articulated bodies (\eg, pulleys). 
We consciously include both physical dynamics reasoning (\eg, interactions between fluids, soft bodies, and rigid bodies), and physical parameter or concept reasoning (\eg, density for fluids; tension, elasticity for soft bodies; mass for rigid bodies).

For instance, our rope and pulley scenarios involve elements of rope, rigid bodies, and articulated bodies; the fluid scenario includes liquids; the cloth scenario covers both cloth and rigid bodies; and the ball scenario focuses on soft balls.
This extensive coverage ensures our dataset provides a comprehensive understanding of the interactions and couplings within these various types of continua, capturing the complexity and diversity of real-world physical phenomena.

In our paper, we focus predominantly on fluids and soft bodies, which are often overlooked in previous works. However, our dataset comprehensively encompasses rigid bodies in all scenarios and articulated bodies (\eg, in the rope scenario). This inclusion leads to our utilization of the continuum concept, enhancing the breadth and relevance of our study.

\section{More Implementation Details}
\subsection{Foundation Models Evaluation Details}
To evaluate currently well-known foundation models such as Gemini (model name: "gemini-pro-vision") and GPT4-V (model name: "gpt-4-vision-preview"), we down-sampled each video to 10 frames and designed specific prompts for different groups of questions. To be concrete, we list our specific prompts in Table~\ref{tb:prompts_fdm}. The rest of the evaluation steps are the same as other baselines.

\subsection{Oracle Model \model Details}\label{sec:ContproDetails}

\textbf{For Code LLM models}, we have tested GPT-4 (gpt-4-0125-preview). We provide full API in Listing~\ref{listing}. We list our prompt in Table~\ref{tb:oracle_prompt}. Examples can refer to Section~\ref{app:q_examples}.

\textbf{For the Visual Perception Module}, we utilize Mask R-CNN (ResNet-R101-FPN) architecture based on Detectron2~\cite{wu2019detectron2}. We use the default config from Detectron2, while the number of classes is different across scenarios. Specifically, the batch size is $16$ for 8 GPUs thus each mini-batch has 2 images per GPU. We train the model for $50k$ iterations, with a learning rate of $0.02$. The proposal number is $1000$ per image. For image size, we keep the original Full HD ($1920\times1080$) resolution.

\textbf{For the Physical Simulation Module}, we adopt MPM for the ball and fluid scenarios respectively, and DPI-Net for the rope and cloth scenarios. We describe the parameter setting for each scenario below.

\textbf{For the fluid scenario}, the physical inference model configurations are listed as follows. Simulations are conducted in a 2D space for efficiency, and the entire scene is rescaled into a square with $x \in [-0.1, 0.1]$, and $y \in [1.0, 1.2]$. Thirty or one hundred points are resampled for each branch of fluid flow, depending on the query conditions. The video frame time step is $1/60$, and the simulation time step is $1/3000$. Initial physical properties include $\kappa = 1 \times 10^3$, default viscosity $\mu = 0.01$, and default density $\rho = 1000$. Learning rates for viscosity $\mu$ and density $\rho$ are $0.001$ and $0.1$ respectively ( under logarithmic density). Gravity $g$ is set as $-0.4$. The property inference stage starts from frame $190$ to the end (frame $250$). MPM grid unit size is $0.0008$. Taichi Snodes CUDA chunk size is $10$, and particle chunk size is $2^{10}$.

\textbf{For the ball scenario}, the physical inference model configurations are similar to the fluid scenario. Simulations are conducted in 2D space with rescaled dimensions. Two hundred points are resampled for each ball, and the von Mises formula is used to model the material. The video frame time step is $1/60$, and the simulation time step is $1/6000/32$ for very high precision to catch up with the ball collision speed in the video. Initial physical properties include default Young's modulus $E = 0.1$, default Poisson's ratio $\nu = 0.1$, and default yield stress $3 \times 10^{-2}$. Learning rates for Young's modulus $E$, Poisson's ratio $\nu$, and yield stress are $0.1$, $0.01$, and $0.1$ respectively. Gravity $g$ is set as $-0.4$, and the friction rate between rigid bodies and balls is $0$. The property inference stage starts from the video start time to the first collision time. The iteration epoch number is $6$. Chamfer loss is used to compare the predicted particles with the ground truth. MPM grid unit size is $0.0016$. Taichi Snodes CUDA chunk size is $100$. Particle chunk size is $2^{10}$.

\textbf{For the rope scenario}, we add object mass as the property in GNN training, which will add attribute relations between nodes. We also separate the soft bodies and rigid bodies by different material relations. The fps of our video is $30$. Other configurations are as follows. The state dimension is $6$ for $x, y, z$, and their speed. We do not use any historical information about the frame for a fair comparison. We set the multi-stage propagation time at $4$. We have trained the model for $50k$ iterations with a batch size of $1$ and a learning rate of $0.0001$. In the inference period, the simulation will start at frame $0$ and predict $30$ frames. For counterfactual and goal-driven simulation, we revise the input mass property, so the attribute relations and simulation will differ.

\textbf{For the cloth scenario}, we also add object mass as the property in GNN training. The parameter setting is similar to the rope scenario. We add a floor to the simulation to represent the table in the scene. In the inference period, the simulation will start at frame $15$ as the first $15$ frames are designed for object observation in which objects will not move. We predict $115$ frames after the 15th frame input. The movement of the clothes is set as the ground truth instead of prediction since the action of objects is caused by the external force of cloth movement.

\section{Experiments of More Baselines}
\subsection{Experiments of Multi-modalities}
We test the performance of CNN-LSTM and MAC with different modalities. We experiment with point cloud features. First, we utilize ULIP-2~\cite{xue2022ulip, xue2023ulip2} pre-trained models with PointBert~\cite{yu2021pointbert} backbones to extract features for all object point clouds in the scenarios. These features are then concatenated together with the vision input, and are fed into vision baselines. Results are shown in Table~\ref{tab:pcresults}. With the help of point clouds, vision models are exposed to large improvements in almost all settings. We articulate that point cloud features can improve vision model performance, providing additional information like object locations and spatial relationships, which is important to predict objects' dynamics.

\subsection{Experiments of More Prompting Methods}\label{app:exp_prompt}
We also tested the performance of MLLMs on different prompting methods such as scenario-specific guidelines, in-context examples, 
and human-explained examples. The prompt examples are shown in Table~\ref{tb:guidelines}, ~\ref{tb:prompt_rephrase}, and Listing ~\ref{in-context}, ~\ref{human-explained}. Results can be found in Table~\ref{tab:apdx_prompting_rslts}.

From method (a) to (i) (check table headers), we draw the average, maximum, and minimum values of various prompting method scores on a radar chart (Figure~\ref{fig:mllm_radar}). For reference, human performance on each question type is plotted as well. For the normalization of visual effects, values on the chart are processed by subtracting the random choice scores.

\section{Qualitative Examples}\label{app:q_examples}
In this section, we show the qualitative examples of the generated programs, which is Python style code, via Code LLM of our \model. As mentioned before, the full API is in Listing~\ref{listing}. We list our prompt in Table~\ref{tb:oracle_prompt}. For each scenario and question type, we show one case. Results are listed in Listing~\ref{eg1}, ~\ref{eg2}, ~\ref{eg3}, ~\ref{eg4}. For better visualization and clarity, we remove some comments, spaces, and blank lines. 

\newpage

\begin{table}
    \caption{Physical reasoning of different modalities. We compare the performance of CNN-LSTM and MAC, w/ and w/o point cloud features.}
 \vspace{0.1in}
    \resizebox{0.99\linewidth}{!}{
    \centering
    \setlength{\tabcolsep}{2pt}
    \renewcommand{\arraystretch}{1.2}
	\begin{tabular}{l|l|cc|cc}
            
            \shline
           Subset & Settings & CNN-LSTM & +Point Cloud & MAC & +Point Cloud \\
            \shline
            \multirow{5}{*}{Rope} & Prop. & 52.7 & 55.0 & 53.3 & 57.7 \\
             & C-Opt. & 74.0 & 75.4 & 74.2 & 76.0 \\
             & C-Ques. & 45.0 & 45.5 & 39.8 & 45.5 \\
             & G-Opt. & 51.2 & 53.8 & 50.3 & 51.7 \\
             & G-Ques. & 6.7 & 10.1 & 6.7 & 5.6 \\
            \shline
            \multirow{7}{*}{Fluid} &Prop. & 54.0 & 55.3 & 30.0 & 50.7 \\
             & C-Opt. & 55.0 & 55.4 & 56.5 & 57.4 \\
             & C-Ques. & 8.6 & 9.5 & 6.9 & 7.8 \\
             & G-Opt. & 57.3 & 58.1 & 51.2 & 58.5 \\
             & G-Ques. & 22.5 & 27.5 & 17.5 & 25.0 \\
             & P-Opt. & 51.4 & 53.2 & 53.5 & 51.9 \\
             & P-Ques. & 12.5 & 10.6 & 12.5 & 13.5 \\
            \shline
            \multirow{3}{*}{Cloth} &Prop. & 46.7 & 47.3 & 59.3 & 59.3 \\
             & P-Opt. & 67.5 & 68.3 & 57.9 & 60.8 \\
             & P-Ques. & 57.3 & 61.7 & 50.7 & 53.3 \\
            \shline
            \multirow{7}{*}{Ball} & Prop. & 54.7 & 55.3 & 48.0 & 52.7 \\
             & C-Opt. & 64.2 & 66.9 & 66.1 & 66.4 \\
             & C-Ques. & 41.8 & 47.5 & 3.3 & 45.9 \\
             & G-Opt. & 54.1 & 60.4 & 58.1 & 52.6  \\
             & G-Ques. & 20.0 & 36.7 & 18.9 & 21.1 \\
             & P-Opt. & 67.4 & 71.2 & 64.4 & 70.5 \\
             & P-Ques. & 45.5 & 53.4 & 46.6 & 55.7 \\
            \shline
	\end{tabular}
	}
	\label{tab:pcresults}
 \vspace{-12pt}
\end{table}

\begin{figure}
\vspace{-0.1in}
\begin{center}
\centerline{\includegraphics[width=.98\linewidth]{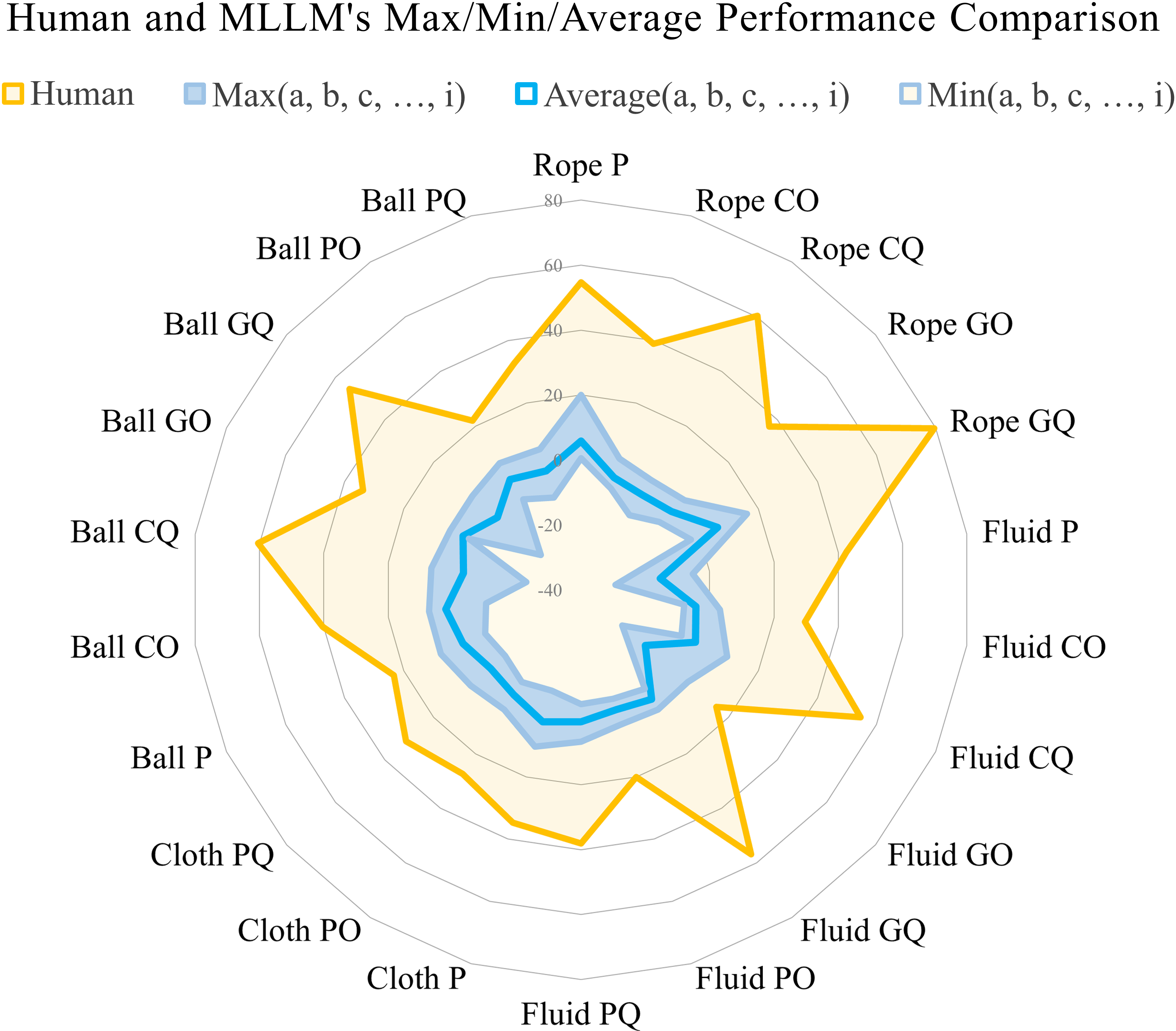}}
\caption{Radar chart of MLLM prompting results on various \dataset subtasks.}
\label{fig:mllm_radar}
\end{center}
\end{figure}

%%%%%

\clearpage

\newlength{\lenx} 
\newlength{\leny} 
\newlength{\lenz} 
\settowidth{\lenx}{\textbf{Q}}
\settowidth{\leny}{\textbf{\emph{Step}}}
\settowidth{\lenz}{\textbf{\emph{E.g.}}}

\begin{table*}
    \small
    \caption{{Question templates, examples, and logical steps in Fluid.}}
    \label{tab:fluid_q}
    \vskip 0.15in
    \centering
    \begin{tabular}{cccl}
    \toprule
    Class & Type & Step Num & Template, Example and Logical Step \\
    \midrule
    \multirow{3}{*}{Density}       &\multirow{3}{*}{Factual}  & \multirow{3}{*}{3}                    & \textbf{Q\ }\hspace{\dimexpr\leny-\lenx} Is the fluid density of the  \_CLR\_ fluid \_CMP\_  that of the \_CLR\_ fluid? \\
     &&& \textbf{\emph{E.g.\ }}\hspace{\dimexpr\leny-\lenz}  Is the fluid density of the pink fluid greater than that of the light blue fluid? \\
    &&&\textbf{\emph{Step\ }} Filter pink fluid. $\rightarrow$ Filter blue fluid. $\rightarrow$ Compare density.  \\
        \midrule
    \multirow{2}{*}{Stick Number}& \multirow{2}{*}{Factual} & \multirow{2}{*}{2} &\textbf{Q\ }\hspace{\dimexpr\leny-\lenx}  How many sticks are there in the video? \\&&&\textbf{\emph{Step\ }} Filter sticks. $\rightarrow$ Count sticks. \\
    
        \midrule
        
    \multirow{3}{*}{Pass}&\multirow{3}{*}{Predictive} & \multirow{3}{*}{3} & \textbf{Q\ }\hspace{\dimexpr\leny-\lenx} Which stick will the fluid from the other \_CLR\_ emitter pass? \\
       &&& \textbf{\emph{E.g.\ }}\hspace{\dimexpr\leny-\lenz}  Which stick will the fluid from the other blue emitter pass?
       \\&&&\textbf{\emph{Step\ }} Filter emitter. $\rightarrow$ Filter sticks. $\rightarrow$ Predict fluid. \\
    
        \midrule
    
        \multirow{3}{*}{Container}&\multirow{3}{*}{Predictive} & \multirow{3}{*}{3} & \textbf{Q\ }\hspace{\dimexpr\leny-\lenx} Which container will fluid from the other \_CLR\_ emitter flow into? \\
       &&& \textbf{\emph{E.g.\ }}\hspace{\dimexpr\leny-\lenz}  Which container will fluid from the other blue emitter flow into?  \\
       &&& \textbf{\emph{Step\ }} Filter emitter. $\rightarrow$ Filter containers. $\rightarrow$ Predict fluid. \\
    
        \midrule
    
        \multirow{3}{*}{Pass}&\multirow{3}{*}{Counterfactual} &\multirow{3}{*}{3}& \textbf{Q\ }\hspace{\dimexpr\leny-\lenx} If \_CLR\_ stick were removed, which stick would \_CLR\_ fluid  pass? \\
       &&& \textbf{\emph{E.g.\ }}\hspace{\dimexpr\leny-\lenz}  If brown stick were removed, which stick would pink fluid pass? \\
       &&& \textbf{\emph{Step\ }} Filter brown stick. $\rightarrow$ Filter pink fluid. $\rightarrow$ Simulate stick removal. \\
    
        \midrule
    
        \multirow{5}{*}{Container}&\multirow{5}{*}{Counterfactual} & \multirow{5}{*}{3} & \textbf{Q\ }\hspace{\dimexpr\leny-\lenx} If the \_CLR\_ stick were removed, which container would \_CLR\_ fluid \\
       &&& \hspace{\dimexpr\leny} flow into? \\
       &&& \textbf{\emph{E.g.\ }}\hspace{\dimexpr\leny-\lenz} If the brown stick were removed, which container would pink fluid \\
       &&& \hspace{\dimexpr\leny} flow into? \\
       &&& \textbf{\emph{Step\ }} Filter brown stick. $\rightarrow$ Filter pink fluid. $\rightarrow$ Simulate stick removal. \\
    
        \midrule
    
        \multirow{3}{*}{Container}&\multirow{3}{*}{Goal-Driven} & \multirow{3}{*}{3} & \textbf{Q\ }\hspace{\dimexpr\leny-\lenx} What can we do to let most of the \_CLR\_ fluid enter the \_CLR\_ container? \\
       &&& \textbf{\emph{E.g.\ }}\hspace{\dimexpr\leny-\lenz} What can we do to let most of the pink fluid enter the gray container?  \\
       &&& \textbf{\emph{Step\ }} Filter gray container. $\rightarrow$ Filter pink fluid. $\rightarrow$ Simulate stick removal. \\
    
    \bottomrule
    \end{tabular}
\end{table*}

\begin{table*}
    \small
    \caption{{Question templates, examples, and logical steps in Rope.}}
    \label{tab:rope_q}
    \vskip 0.15in
        % \resizebox{1\linewidth}{!}{
    \centering
    \begin{tabular}{cccl}
    \toprule
    Class & Type & Step Num&  Template, Example and Logical Steps \\
    \midrule
    \multirow{3}{*}{Shape}       &\multirow{3}{*}{Factual}  &    \multirow{3}{*}{3}    & \textbf{Q\ }\hspace{\dimexpr\leny-\lenx} How many \_SHP\_ \_OBJ\_s are there in the video?\\
        &&& \textbf{\emph{E.g.\ }}\hspace{\dimexpr\leny-\lenz}  How many solid pulleys are there in the video? \\
        &&& \textbf{\emph{Step\ }} Filter pulleys. $\rightarrow$ Filter solid objects. $\rightarrow$ Count objects. \\

        \midrule
    \multirow{3}{*}{Color}&\multirow{3}{*}{Factual}&\multirow{3}{*}{2} & \textbf{Q\ }\hspace{\dimexpr\leny-\lenx}  How many \_CLR\_ objects are there in the video?\\
        &&& \textbf{\emph{E.g.\ }}\hspace{\dimexpr\leny-\lenz}  How many blue objects are there in the video?\\
        &&& \textbf{\emph{Step\ }} Filter blue objects. $\rightarrow$ Count objects. \\
    
        \midrule
    \multirow{3}{*}{Existence}&\multirow{3}{*}{Factual} &\multirow{3}{*}{2} & \textbf{Q\ }\hspace{\dimexpr\leny-\lenx} Is there any \_OBJ\_ in the video?\\
        &&& \textbf{\emph{E.g.\ }}\hspace{\dimexpr\leny-\lenz}   Is there any blue cube in the video?\\
            &&& \textbf{\emph{Step\ }} Filter blue cube. $\rightarrow$ Check existence. \\
        
            \midrule

            \multirow{3}{*}{Mass} &\multirow{3}{*}{Factual} & \multirow{3}{*}{3} & \textbf{Q\ }\hspace{\dimexpr\leny-\lenx}  Is the mass of the \_OBJ\_ \_CMP\_ \_FAC\_ that of the \_OBJ\_?\\
        &&&  \textbf{\emph{E.g.\ }}\hspace{\dimexpr\leny-\lenz}    Is the mass of the blue sphere greater than half that of the green cube? \\
        &&& \textbf{\emph{Step\ }} Filter blue sphere. $\rightarrow$ Filter green cube. $\rightarrow$ Compare mass. \\

        \midrule
    
    \multirow{3}{*}{Tension} &\multirow{3}{*}{Factual}&\multirow{3}{*}{3} & \textbf{Q\ }\hspace{\dimexpr\leny-\lenx}  Is the tension of the \_CLR\_ rope \_CMP\_ \_FAC\_ that of the \_CLR\_ rope?\\
        &&&  \textbf{\emph{E.g.\ }}\hspace{\dimexpr\leny-\lenz}   Is the tension of 
    the blue rope greater than half that of the green rope? \\
        &&& \textbf{\emph{Step\ }} Filter blue rope. $\rightarrow$ Filter green rope. $\rightarrow$ Compare tension. \\
    
        \midrule
    \multirow{3}{*}{Rotation} &\multirow{3}{*}{Counterfactual}&\multirow{3}{*}{3} &  \textbf{Q\ }\hspace{\dimexpr\leny-\lenx}  If the \_OBJ\_ were heavier, which direction would the \_OBJ\_ move?
    \\
        &&&   \textbf{\emph{E.g.\ }}\hspace{\dimexpr\leny-\lenz}   If the blue sphere were heavier, which direction would the green cube move? \\
        &&& \textbf{\emph{Step\ }} Filter blue sphere. $\rightarrow$ Filter green cube. $\rightarrow$ Simulate mass change. \\
    
        \midrule
    \multirow{3}{*}{Direction}   &\multirow{3}{*}{Counterfactual} &\multirow{3}{*}{3}  &  \textbf{Q\ }\hspace{\dimexpr\leny-\lenx}  If the \_OBJ\_ were heavier, which direction would the \_OBJ\_ move? \\
        &&&   \textbf{\emph{E.g.\ }}\hspace{\dimexpr\leny-\lenz}  If the blue cube were heavier, which direction would the brown sphere move? \\
        &&& \textbf{\emph{Step\ }} Filter blue cube. $\rightarrow$ Filter brown sphere. $\rightarrow$ Simulate mass change. \\
    
        \midrule
    \multirow{3}{*}{Mass Goal}      &\multirow{3}{*}{Goal-Driven}&\multirow{3}{*}{3}   & \textbf{Q\ }\hspace{\dimexpr\leny-\lenx}   If we want the \_OBJ\_ to \_ENT\_, what can we do? \\ 
    &&&   \textbf{\emph{E.g.\ }}\hspace{\dimexpr\leny-\lenz}   If we want the yellow cube to move up, what can we do? \\ 
        &&& \textbf{\emph{Step\ }} Filter yellow cube. $\rightarrow$ Simulate mass change. $\rightarrow$ Filter motion or direction. \\

    \bottomrule
    \end{tabular}
\end{table*}

\begin{table*}
    \small
    \caption{{Question templates, examples, and logical steps in Cloth.}}
    \label{tab:cloth_q}
    \vskip 0.15in
    \centering
    \begin{tabular}{cccl}
    \toprule
    Class & Type & Step Num & Template, Example and Logical Steps \\
    
    \midrule
    \multirow{3}{*}{Elasticity}       &\multirow{3}{*}{Factual}                      &\multirow{3}{*}{3}& \textbf{Q\ }\hspace{\dimexpr\leny-\lenx} Is the elasticity of the \_POS\_ cloth much \_CMP\_  that of the other?  \\
        &&& \textbf{\emph{E.g.\ }}\hspace{\dimexpr\leny-\lenz} Is the elasticity of the left cloth much greater than that of the other? \\
        &&& \textbf{\emph{Step\ }} Filter left cloth. $\rightarrow$ Filter right cloth. $\rightarrow$ Compare elasticity. \\

        \midrule
    \multirow{3}{*}{Bending}       &\multirow{3}{*}{Factual}                      &\multirow{3}{*}{3}& \textbf{Q\ }\hspace{\dimexpr\leny-\lenx} Is the \_POS\_ cloth much \_CMP\_ to bend or have wrinkles than the other? \\
        &&& \textbf{\emph{E.g.\ }}\hspace{\dimexpr\leny-\lenz} Is the right cloth much harder to bend or have wrinkles than the other? \\
        &&& \textbf{\emph{Step\ }} Filter left cloth. $\rightarrow$ Filter right cloth. $\rightarrow$ Compare bending. \\
    
        \midrule
    \multirow{3}{*}{Fall Over}       &\multirow{3}{*}{Predictive}                      &\multirow{3}{*}{2}& \textbf{Q\ }\hspace{\dimexpr\leny-\lenx} Does the \_CLR\_ \_OBJ\_ fall over? \\
        &&& \textbf{\emph{E.g.\ }}\hspace{\dimexpr\leny-\lenz} Does the green plate fall over? \\
        &&& \textbf{\emph{Step\ }} Filter green plate. $\rightarrow$ Predict fall over. \\
    
        \midrule
    \multirow{3}{*}{Collision}       &\multirow{3}{*}{Predictive}                      &\multirow{3}{*}{3}& \textbf{Q\ }\hspace{\dimexpr\leny-\lenx} Does the \_CLR\_ \_OBJ\_ collide with the \_CLR\_ \_OBJ\_? \\
        &&& \textbf{\emph{E.g.\ }}\hspace{\dimexpr\leny-\lenz} Does the green plate collide with the gray pillar?  \\
        &&& \textbf{\emph{Step\ }} Filter green plate. $\rightarrow$ Filter gray pillar. $\rightarrow$ Predict collision. \\
    
        \midrule
    \multirow{3}{*}{Touch}       &\multirow{3}{*}{Predictive}                      &\multirow{3}{*}{3}& \textbf{Q\ }\hspace{\dimexpr\leny-\lenx} Is the \_CLR\_ \_OBJ\_ finally in touch with the \_CLR\_ \_OBJ\_?  \\
        &&& \textbf{\emph{E.g.\ }}\hspace{\dimexpr\leny-\lenz} Is the green plate finally in touch with the gray pillar? \\
        &&& \textbf{\emph{Step\ }} Filter green plate. $\rightarrow$ Filter gray pillar. $\rightarrow$ Predict touch. \\
    
        \midrule
    \multirow{3}{*}{Pose}       &\multirow{3}{*}{Predictive}    &\multirow{3}{*}{2}& \textbf{Q\ }\hspace{\dimexpr\leny-\lenx} Which phrase below can best describe the final pose of the \_CLR\_ \_OBJ\_?  \\
        &&& \textbf{\emph{E.g.\ }}\hspace{\dimexpr\leny-\lenz} Which phrase below can best describe the final pose of the green plate?  \\
        &&& \textbf{\emph{Step\ }} Filter green plate. $\rightarrow$ Predict pose. \\
    
    \bottomrule
    \end{tabular}
    % }
\end{table*}

\begin{table*}
    \small
    \caption{{Question templates, examples, and logical steps in Ball.}}
    \label{tab:ball_q}
    \vskip 0.15in
    \centering
    \begin{tabular}{cccl}
    \toprule
    Class & Type & Step Num & Template, Example and Logical Steps \\
    
    \midrule
    \multirow{3}{*}{Elasticity}       &\multirow{3}{*}{Factual}                      &\multirow{3}{*}{3}& \textbf{Q\ }\hspace{\dimexpr\leny-\lenx} Is the elasticity of the \_CLR\_ ball much \_CMP\_  the  \_CLR\_  ball?  \\
     &&& \textbf{\emph{E.g.\ }}\hspace{\dimexpr\leny-\lenz} Is the elasticity of the brown ball much greater than the purple ball? \\
        &&& \textbf{\emph{Step\ }} Filter brown ball. $\rightarrow$ Filter purple ball. $\rightarrow$ Compare elasticity. \\

    \midrule
    \multirow{3}{*}{Plasticity}       &\multirow{3}{*}{Factual}                      &\multirow{3}{*}{3}& \textbf{Q\ }\hspace{\dimexpr\leny-\lenx} Is the plasticity of the \_CLR\_ ball much \_CMP\_  the  \_CLR\_  ball?  \\
     &&& \textbf{\emph{E.g.\ }}\hspace{\dimexpr\leny-\lenz} Is the plasticity of the brown ball much greater than the purple ball? \\
        &&& \textbf{\emph{Step\ }} Filter brown ball. $\rightarrow$ Filter purple ball. $\rightarrow$ Compare plasticity. \\

     \midrule
    \multirow{3}{*}{Final Drop}       &\multirow{3}{*}{Predictive}                      &\multirow{3}{*}{3}& \textbf{Q\ }\hspace{\dimexpr\leny-\lenx} Will the \_CLR\_ ball finally drop into the \_POS\_ pit? \\
     &&& \textbf{\emph{E.g.\ }}\hspace{\dimexpr\leny-\lenz} Will the brown ball finally drop into the left pit? \\
        &&& \textbf{\emph{Step\ }} Filter brown ball. $\rightarrow$ Filter left pit. $\rightarrow$ Predict final drop. \\
    
     \midrule
    \multirow{5}{*}{Remove}       &\multirow{5}{*}{Counterfactual}                      &\multirow{5}{*}{3}& \textbf{Q\ }\hspace{\dimexpr\leny-\lenx} If we removed the \_CLR\_ floating wall and other balls,  which pit would 
    \\&&&\hspace{\leny} \ the \_CLR\_ ball drop into?  \\
     &&& \textbf{\emph{E.g.\ }}\hspace{\dimexpr\leny-\lenz} If we removed the yellow floating wall and other balls,  which pit would 
     \\&&&\hspace{\leny} \ the brown ball drop into?  \\
        &&& \textbf{\emph{Step\ }} Filter yellow floating wall. $\rightarrow$ Filter brown ball. $\rightarrow$ Simulate removal. \\
    
     \midrule
    \multirow{3}{*}{Drop}   &  \multirow{3}{*}{Goal-Driven}     &\multirow{3}{*}{3}& \textbf{Q\ }\hspace{\dimexpr\leny-\lenx} What can we do to make the \_CLR\_ ball drop into the  \_POS\_ pit?  \\
     &&& \textbf{\emph{E.g.\ }}\hspace{\dimexpr\leny-\lenz} What can we do to make the pink ball drop into the right pit? \\
        &&& \textbf{\emph{Step\ }} Filter pink ball. $\rightarrow$ Simulate removal. $\rightarrow$ Filter right pit. \\

    \bottomrule
    \end{tabular}
    % }
    \end{table*}
    
\begin{figure*}[ht]
  \centerline{\includegraphics[width=.95\linewidth]{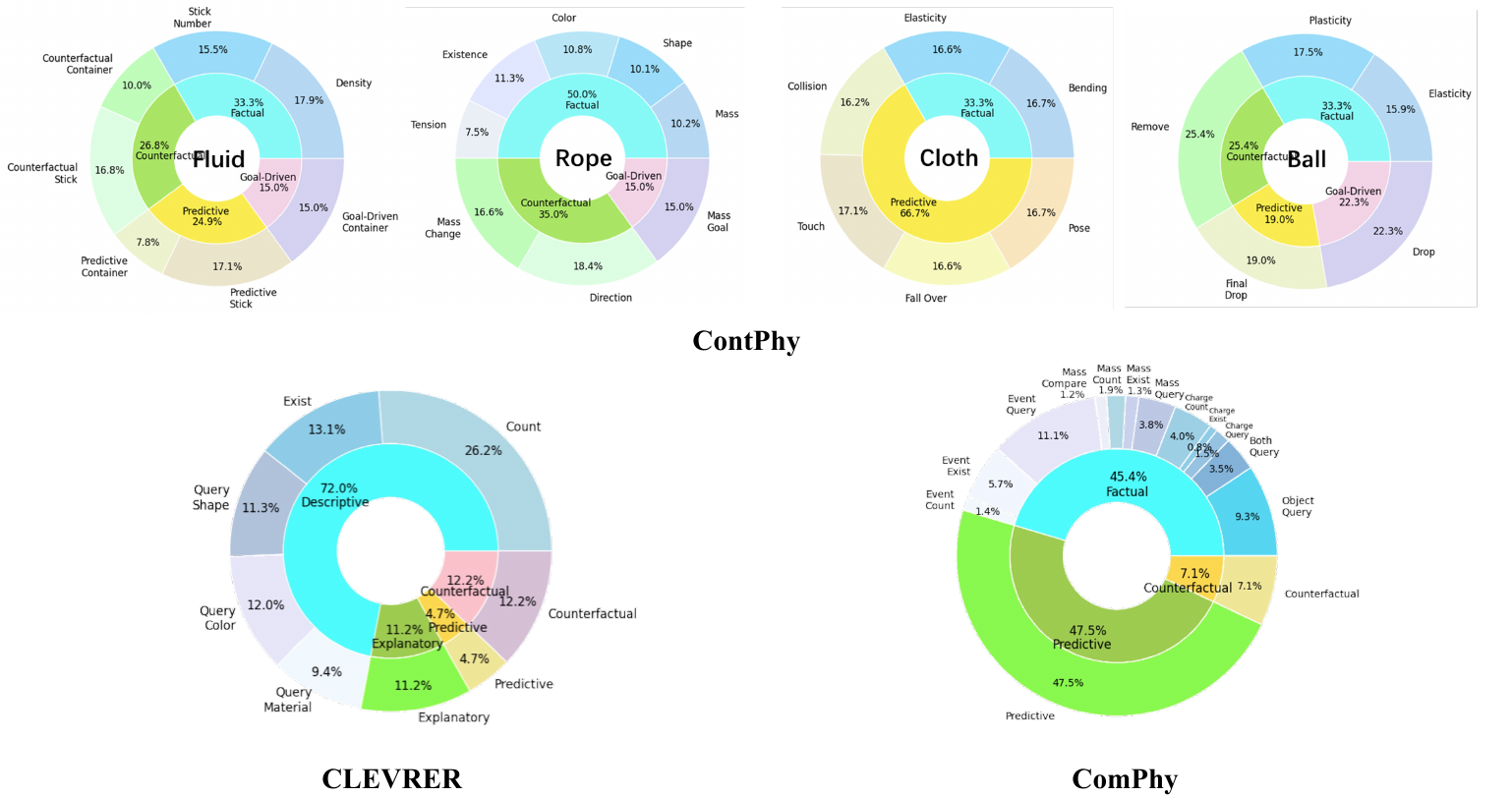}}
  \caption{Question type distribution of ContPhy, and two related datasets.}
  \label{fig:contphy_qtype_pie}
\end{figure*}

\begin{figure*}
  \centering
  \centerline{\includegraphics[width=.92\linewidth]{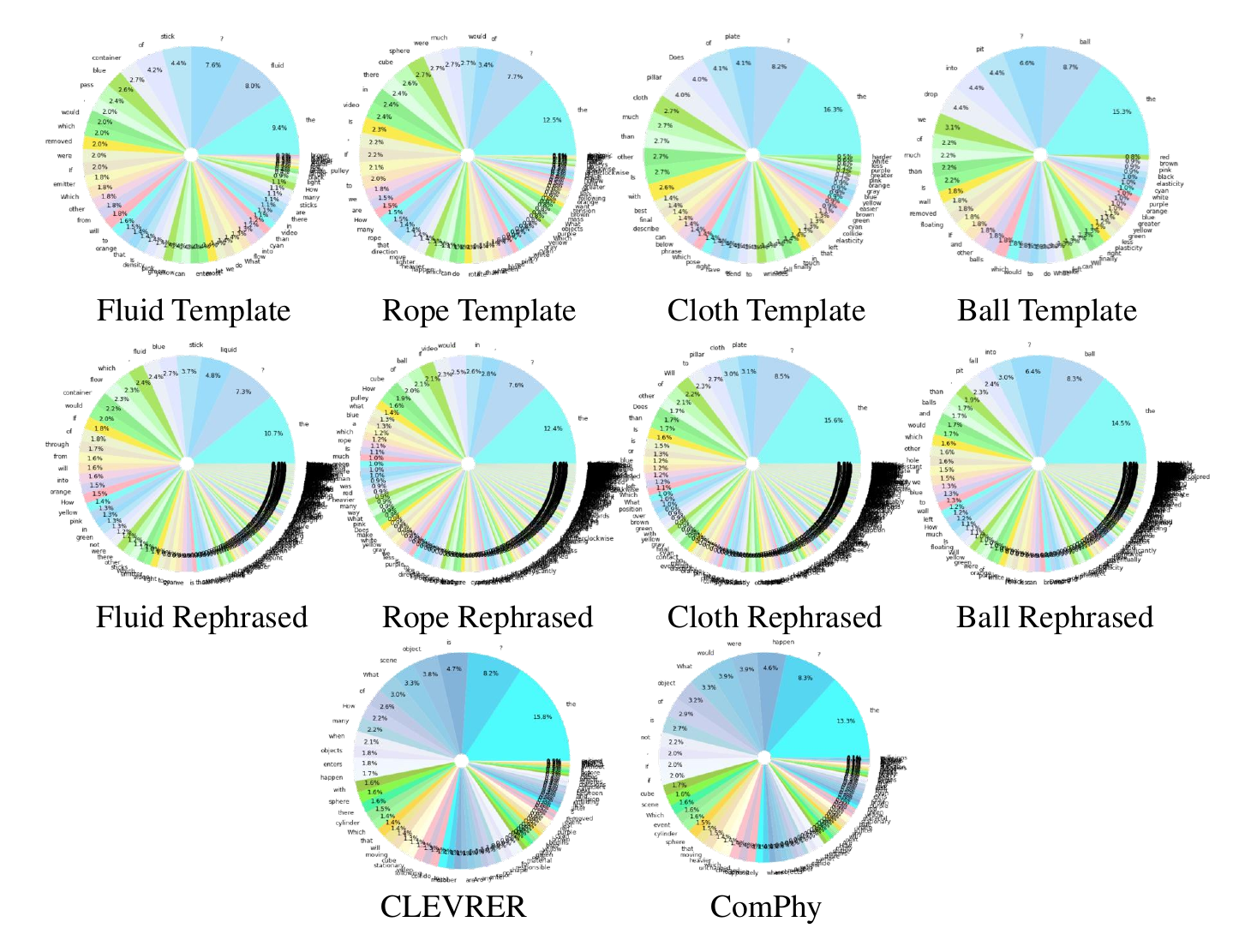}}

  \caption{Word distribution of ContPhy, and two related datasets.}
  \label{fig:contphy_wdist_pie}
\end{figure*}

\begin{table*}
    \small
    \caption{Full question statistics and comparison w/ and w/o LLM rephrase.}
    \vskip 0.15in
    \centering
    \label{tab:apdx_stat}
\begin{tabular}{ccccccc}
    \toprule
        Scenario  & Question Types & Generation Method & TTR & Len Avg & Len Var & F-K Grade Level \\ 
        \midrule

        \multirow{2}{*}{Fluid} & \multirow{2}{*}{7 Types} &  Template & 0.0096  & 13.1 & 3.9 & 4.4 \\ \cmidrule(lr){3-7}

        & & +LLM Rephrase & 0.052  & 13.6 & 10.7 & 4.5 \\ 
        \midrule

        \multirow{2}{*}{Rope} & \multirow{2}{*}{8 Types} &  Template & 0.0096  & 13.0 & 6.9 & 3.1 \\ \cmidrule(lr){3-7}

        & & +LLM Rephrase & 0.053  & 13.0 & 11.3 & 3.1 \\ 
        \midrule

        \multirow{2}{*}{Cloth} & \multirow{2}{*}{6 Types} &  Template & 0.0089  & 12.2 & 8.5 & 4.1 \\ \cmidrule(lr){3-7}

        & & +LLM Rephrase & 0.068  & 11.7 & 11.4 & 3.9 \\ 
        \midrule

        \multirow{2}{*}{Ball} & \multirow{2}{*}{5 Types} &  Template & 0.0066 & 15.2 & 10.2 & 4.0 \\ \cmidrule(lr){3-7}

        & & +LLM Rephrase & 0.049  & 15.6 & 19.2 & 4.1 \\ 
        \midrule
        
        ComPhy~\cite{chen2022comphy} & 14 Types & - & 0.0005  & 12.0 & 8.7 & 4.0 \\  \midrule

        CLEVRER~\cite{yi2019clevrer} & 8 Types & - & 0.00008  & 12.2 & 12.6 & 5.3 \\ 

        \bottomrule
        
    \end{tabular}
    % }
    \end{table*}

\begin{table*}[t]
\caption{{Detailed illustration of prompts that we use to evaluate Gemini and GPT4-V.}}
\label{tb:prompts_fdm}
\vskip 0.15in
\centering
\begin{tabular}{ll}
\toprule
Settings& Texts\\
\midrule
General Prompt& For now I am giving you a set of frames extracted from a video,\\
 & with some questions related to the video. You need to answer \\
 &questions in the given order. For each question please answer it in a \\
 &fixed format following the comment after the question. For the overall \\
 &output, you need to list the answers for all questions in the original \\
 &question order, and divide them by ``;". For example, if you have \\
 &two questions, and the answers are ``A B C" and ``yes", then you need to \\
 &respond with ``A B C;yes". Please do NOT add any other text in \\
 &your response. Thank you!\\ 
    \midrule
 Multiple Choice& Please answer with all correct choices listed in alphabet order, divided \\
 &by spaces. For example, you can respond with ``A B C". Please do NOT \\
 &add any other text in your response.\\
    \midrule
Single Choice& Please answer with the correct choice. For example, you can \\
 &respond with ``A". Please do NOT add any other text in your response.\\
  \midrule
Open-Ended (Number Answer)& Please answer a number. For example, you can respond with \\
 &``3". Please do NOT add any other text in your response.\\
\midrule
Open-ended (Yes or No)& Please answer with ``yes" or ``no". For example, you can \\
 &respond with ``yes". Please do NOT add any other text in your \\
 &response.\\
       \midrule
Open-ended (Yes, No, or Can not Answer)& Please answer with ``yes", ``no", or ``can not answer". For example, \\
 &you can respond with ``can not answer". Please do NOT add \\
 &any other text in your response.\\
 \bottomrule
\end{tabular}
\vskip -0.1in
\end{table*}

\begin{table*}[t]
    \caption{Results of experiments across different MLLM prompting methods.}
 \vspace{0.1in}
    \resizebox{1\linewidth}{!}{
    \centering
    \setlength{\tabcolsep}{4pt}
    \renewcommand{\arraystretch}{1.2}
	\begin{tabular}{l|l|cccccccccc|c|c}
        \shline
        \multirow{2}{*}{Subset} & \multirow{2}{*}{Settings} & \multicolumn{10}{c|}{Different MLLM Prompting Methods} & \multirow{2}{*}{Random} & \multirow{2}{*}{Human} \\
       & & (a) & (b) & (c) & (d) & (e) & (f) & (g) & (h) & (i) & Average & &  \\
        \shline
        \multirow{5}{*}{Rope} & Prop. & 35.5 & 34.0 & 33.5 & 34.5 & 39.0 & 34.0 & 30.5 & 32.0 & 50.0 & 35.9  & 30.0 & 84.7 \\
        & C-Opt. & 48.2 & 44.4 & 46.6 & 51.2 & 53.4 & 46.6 & 43.8 & 44.0 & 47.8 & 47.3  & 51.3 & 90.2 \\
        & C-Ques. & 12.0 & 5.6 & 14.8 & 13.4 & 12.0 & 11.3 & 9.2 & 5.6 & 2.1 & 9.6  & 14.7 & 75.0 \\
        & G-Opt. & 51.6 & 48.9 & 54.7 & 56.1 & 57.4 & 48.9 & 49.3 & 47.1 & 54.7 & 52.1  & 55.2 & 91.9 \\
        & G-Ques. & 10.3 & 10.3 & 20.7 & 12.1 & 19.0 & 8.6 & 6.9 & 1.7 & 6.9 & 10.7  & 4.5 & 84.0 \\
        \shline
        \multirow{7}{*}{Fluid} & Prop. & 10.0 & 28.0 & 22.0 & 24.0 & 21.0 & 19.0 & 11.0 & 22.0 & 4.0 & 17.9  & 33.3 & 75.8 \\
    & C-Opt. & 47.3 & 48.0 & 45.7 & 48.3 & 46.0 & 46.3 & 45.0 & 55.3 & 56.0 & 48.7  & 52.9 & 82.5 \\
    & C-Ques. & 5.1 & 6.4 & 2.6 & 5.1 & 2.6 & 0.0 & 2.6 & 15.4 & 2.6 & 4.7  & 6.0 & 60.6 \\
    & G-Opt. & 44.4 & 63.3 & 40.8 & 42.6 & 36.7 & 40.8 & 43.2 & 60.4 & 42.0 & 46.0  & 59.9 & 75.0 \\
    & G-Ques. & 11.3 & 11.3 & 5.7 & 7.5 & 3.8 & 5.7 & 5.7 & 11.3 & 5.7 & 7.6  & 7.5 & 64.3 \\
    & P-Opt. & 52.4 & 51.2 & 53.1 & 51.6 & 48.8 & 49.2 & 52.0 & 54.3 & 57.1 & 52.2  & 53.8 & 73.9 \\
    & P-Ques. & 5.8 & 8.7 & 5.8 & 5.8 & 2.9 & 4.3 & 4.3 & 11.6 & 0.0 & 5.5  & 4.8 & 42.9 \\
        \shline
        \multirow{3}{*}{Cloth} &Prop. & 42.0 & 54.0 & 46.0 & 54.0 & 54.0 & 47.0 & 39.0 & 48.0 & 57.0 & 49.0  & 46.7 & 81.4 \\
    & P-Opt. & 50.1 & 56.1 & 50.1 & 45.9 & 50.3 & 47.7 & 50.1 & 55.7 & 49.2 & 50.6  & 52.2 & 79.6 \\
    & P-Ques. & 43.0 & 50.0 & 43.0 & 37.0 & 42.0 & 38.5 & 41.5 & 51.0 & 40.5 & 42.9  & 46.0 & 77.3 \\
        \shline
        \multirow{7}{*}{Ball} & Prop. & 54.0 & 54.0 & 52.0 & 53.0 & 46.0 & 56.0 & 58.0 & 61.0 & 47.0 & 53.4  & 53.5 & 76.9 \\
    & C-Opt. & 60.9 & 60.1 & 56.4 & 47.3 & 43.2 & 60.5 & 57.2 & 57.6 & 58.4 & 55.7  & 53.6 & 93.9 \\
    & C-Ques. & 29.6 & 37.0 & 28.4 & 13.6 & 7.4 & 34.6 & 27.2 & 28.4 & 37.0 & 27.0  & 30.4 & 90.9 \\
    & G-Opt. & 54.1 & 60.1 & 57.9 & 55.2 & 57.4 & 54.1 & 53.9 & 55.6 & 55.2 & 55.9  & 55.9 & 89.7 \\
    & G-Ques. & 24.6 & 34.4 & 31.1 & 6.6 & 11.5 & 23.0 & 27.9 & 32.8 & 26.2 & 24.2  & 30.2 & 84.6 \\
    & P-Opt. & 51.7 & 47.1 & 52.9 & 51.1 & 43.7 & 50.6 & 52.3 & 56.9 & 52.9 & 51.0  & 50.6 & 72.5 \\
    & P-Ques. & 25.9 & 17.2 & 27.6 & 20.7 & 15.5 & 25.9 & 27.6 & 31.0 & 25.9 & 24.1  & 25.9 & 58.8 \\
        \shline
	\end{tabular}
	}
    \label{tab:apdx_prompting_rslts}

    \vspace{20pt}

    \setlength{\tabcolsep}{10pt}
    \renewcommand{\arraystretch}{1.3}
    \centering
    \normalsize
    \begin{tabular}{|c|l|}
        \hline
       \textbf{Notations} & \textbf{Prompt Methods} \\
        \hline
        (a) & Question Only (Visual Input, 0-shot) \\
        (b) & Question Only (Text Only) \\
        (c) & Scenario-Specific Guideline \\
        (d) & In-Context QA Examples \\
        (e) & Human Explained Examples \\
        (f) & Upsampled Video (11→16 Frames, Higher Resolution) \\
        (g) & LLM-rephrased Questions (Visual Input, 0-shot) \\
        (h) & LLM-rephrased Questions (Text Only) \\
        (i) & NEWTON Approach (Text Only)\\
        \hline
    \end{tabular}
\end{table*}

\begin{table*}[t]
    \caption{Prompt example of question rephrasing.}
    \label{tb:prompt_rephrase}
    \vskip 0.15in
    \centering
    \begin{tabular}{l}
    \toprule
    Texts\\
    \midrule
    I am looking for assistance in rephrasing this question.\\\\
    
    My primary goal is to ensure that the essence and meaning of the question, along with the content of each \\
    option, remain unchanged. It is crucial that the sequence of the options is preserved so that the correct \\
    answer corresponds directly with the original question.\\\\
    
    Below, I will provide the question with its options. Please rephrase it as diversely as possible, maintaining \\
    strict adherence to their original meaning. Make questions readable and understandable for common people \\
    as well. Please only return rephrased question (with its rephrased options if it has). Do not add any \\
    other text. Please keep the color name and the object name unchanged. Please do not change the word \\
    \textit{`elastic'/`plastic'} or \textit{`elasticity'/`plasticity'}. If the object name has \textit{`the other'} \\
    description, let this description stay unchanged. If you think the option is too hard to rephrase, you can keep \\
    it unchanged. Also, keep the option format unchanged. \\\\
    
    For example, if I give you the following question: \\\\

    If the gray stick were removed, which stick would orange fluid pass?\\
    A. Pink stick\\
    B. Brown stick\\
    C. Cyan stick\\\\
    
    You may response:\\\\
    
    If the gray stick were not there, which stick would the orange liquid flow through?\\
    A. Pink stick\\
    B. Cyan stick\\
    C. Cyan stick\\\\
    
    PLEASE STRICTLY FOLLOW the above response format. Otherwise, we could not use the program to process your \\
    response. OK, here is the original question you will rephrase.\\\\

\texttt{QUESTIONS\_INSERT\_HERE}\\\\

Thank you for your assistance!\\
     \bottomrule
    \end{tabular}
    \end{table*}

\begin{table*}[t]
    \caption{Prompt example of scenario-specific guidelines.}
    \label{tb:guidelines}
    \vskip 0.15in
    \centering
    \begin{tabular}{ll}
    \toprule
    Settings & Texts\\
    \midrule
    Basic-Prompt & For now I am giving you a set of frames extracted from a video, with some questions related \\ 
    & to the video. You need to answer questions in the given order. For each question please \\
    & answer it in a fixed format following the comment after the question. For the overall output, \\
    & you need to list the answers for all questions in the original question order, and divide them \\
    & by \textit{`;'}. For example, if you have two questions, and the answers are `A B C' and `yes', then \\
    & you need to respond with `A B C;yes'.\\
    & Please do NOT add any other text in your response. Thank you!\\

    \midrule
    Rope& \texttt{BASIC\_PROMPT\_INSERT\_HERE}\\
    & Here is some additional prompts for you. \\
    & Scenario Introduction: An array of pulleys, including both movable and fixed types, along \\
    & with anchor points, is arranged on a wall. Ropes are configured with their ends connected \\
    & to pulleys, loads, or anchor points, and can be wound around the pulleys. These loads \\
    & possess varying masses, interacting with other forces in the system, leading to the emergence \\
    & of distinct motion patterns. $|$ \\
    & The primary objective of the model is to identify the tension distributions within this \\
    & elementary rope system. Additionally, it is tasked with recognizing potential correlations or \\
    & constraints among objects in motion, such as the coordinated movement of loads and the \\
    & rotation of pulleys on a single rope. Moreover, the model is expected to infer numerical \\
    & relationships between the loads' masses.\\

\midrule
    Fluid & \texttt{BASIC\_PROMPT\_INSERT\_HERE}\\
    & Here is some additional prompts for you. \\
    & Scenario Introduction: In this device, various liquids of different densities and viscosities, \\
    & each represented by distinct colors, are released from corresponding emitters situated at the \\
    & uppermost part of the apparatus. Under the influence of gravity, these liquids descend and \\
    & traverse a series of fixed ramps (resembling sticks). This arrangement causes alterations in their \\
    & flow direction. Ultimately, the liquids are funneled into containers at the bottom. This process \\
    & highlights distinctive behaviors arising from the interaction of multiple fluids, attributable to \\
    & their significantly varied densities. Our research is oriented towards formulating inquiries \\
    & pertaining to the physical properties of these liquids and the dynamic trajectories they exhibit. \\

    \midrule
    Cloth & \texttt{BASIC\_PROMPT\_INSERT\_HERE}\\
    & Here are some additional prompts for you. \\
    & Scenario Introduction: A small table hosts an assortment of objects, including pillars and \\
    & plates of varying sizes, colors, and masses. Two square pieces of cloth, each possessing \\
    & distinct stretching, bending characteristics, and frictional properties, are gripped at one edge \\
    & and moved forward to cover these objects, causing possible collision events. Clothes are then \\
    & promptly released. The fabric obstructs the view of the objects but also delineates their shapes \\
    & through its deformable surface. Objects may topple over if they exceed a certain height or \\
    & have low mass, resulting in observable changes in the fabric's dynamic 3D surface geometry. \\
    & This scenario serves as a test for a model's capacity to discern the physical attributes of the \\
    & fabrics and to predict the spatial behavior of the concealed objects in dynamic situations. \\

    \midrule
    Ball & \texttt{BASIC\_PROMPT\_INSERT\_HERE}\\
    & Here are some additional prompts for you. \\
    & Scenario Introduction: A playground contains obstacles of different colors, and poses, along with \\
    & pits randomly arranged within. Soft balls with varying deformation resistance or plasticity yield \\
    & are launched randomly within the space, with varying initial positions. These balls undergo a \\
    & sequence of dynamic movements, including bouncing and permanent deformation. Ultimately,\\
    &  some may collide with obstacles and fall into pits. This experimental scenario serves as a test to \\
    & determine whether the model can accurately discern the elasticity and plasticity properties of the \\
    & soft bodies and moreover make dynamic predictions and inferences based on these observations.
    \\
     \bottomrule
    \end{tabular}
    \end{table*}

\begin{figure*}[t]
\begin{lstlisting}[language=Python, xleftmargin=.03\textwidth, xrightmargin=.03\textwidth,firstnumber=1, caption=Prompt example of in-context prompting., label={in-context}]
# Prompt Example: In-Context Examples

SCENARIO_EXAMPARS = {
    "fluid": # final 17 as the example
        [
            "Here are some additional examples for you to get the feeling of how to solve the problem.",
            [data_dirs["fluid"]["videos"] + "/17/frames/output_Full_ori/frame_00247.png"],
            "Above is the last frame of the example video. The example questions are: (1)\"Is the density of the blue fluid greater than that of the green fluid?\" (2)\"Will the yellow fluid which is emitting at the last frame finally enter the white container?\" The correct answers for these questions are (1)\"no\" (2)\"no\". \n\nOK. Since you have got some examples for reference. The following questions are for you!"
        ],
    ...
}
\end{lstlisting}
\end{figure*}

\begin{figure*}[t]
\begin{lstlisting}[language=Python, xleftmargin=.03\textwidth, xrightmargin=.03\textwidth,firstnumber=1, caption=Prompt example of human explanation., label={human-explained}]
# Prompt Example: Human Explained Examples

  SCENARIO_HUMAN_EXPL = {
    "fluid": # final 17 as the example
        [
            "Here are some additional detailed guidance/tutorials for you to get the feeling of how to solve the problems.",
            [data_dirs["fluid"]["videos"] + "/17/frames/output_Full_ori/frame_00000.png"],
            [data_dirs["fluid"]["videos"] + "/17/frames/output_Full_ori/frame_00247.png"],
            "The above 2 images are the first and the last frames from an example video. In the next question-answering part we will give you some sparsely sampled frames in another similar video. In both the example and target videos, users will first see colored fluids emitted from the top emitters which look like colored cubes. Then the fluids will drop upon and flow down along several colored ramps that look like sticks, and finally, the fluids will enter one or several colored containers at the bottom, which are constructed by several long sticks. Make sure you can detect these key objects. The fluids have different colors, which represent different densities. The fluids will collide with each other and the ramps during the process. Finally, in the container, they will stratify into obvious layers. Note that the lighter fluid will float on the heavier fluid! This is very important when you choose answers about density. Also, note that the process is governed by the gravity. The questions will be about the density, the flow direction, the collision, and the final container of the fluids. At the end of the video, there might be some fluids starting to emit but not yet entering any containers. You need to predict which container and stick they will contact with. \nTake the above 2 example images as an example, in the last frame, you can see blue fluid floating upon the green fluid in the white container, while the yellow fluid is floating upon the green fluid in the gray container. So you can answer the density question based on this observation, which means that most of the time you can only analyze the last frame to determine the density relations. \n Also, in the last frame, you may notice the green fluid is emitting on the left top side. You can predict that, under gravity, it will drop onto the orange stick, flow along the orange stick, and then drop onto the green stick, then flow along the green stick. Then it will finally drop into the white container. Through this reasoning logic chain, you can solve some problems like the following examples. \nThe example questions are: (1)\"Is the density of the blue fluid greater than that of the green fluid?\" (2)\"Will the green fluid which is emitting at the last frame finally enter the gray container?\" The correct answers for these questions are (1)\"no\" (2)\"no\". \n\nOK. Since you have got some examples for reference. The following questions are for you! Forget the above images now and just focus on the following images."
        ],
    ...
}
\end{lstlisting}
\end{figure*}

\begin{table*}[t]
\caption{{Detailed illustration of prompts that we use in program generation of \model oracle model.}}
\label{tb:oracle_prompt}
\vskip 0.15in
\centering
\begin{tabular}{l}
\toprule
Texts\\
\midrule

\texttt{API\_INSERT\_HERE}\\
\\
Write a function using Python and the SoftScene class (above) that could be executed to provide an answer \\to the query. \\
\\
Consider the following guidelines:\\
- Use base Python (comparison, sorting) for basic logical operations, math, etc.\\
- If comparing the property of two objects, call `query\_pair' function.\\
- `query\_single' function can be only used once. \\
- gd (goal-driven) scene tends to start with 'what can we do to ...'.\\
- cf (counterfactual) scene tends to start with 'if we ..., what will happen'.\\
- Only answer with a function starting def execute\_command(video).\\
- Different objects in different scenes:\\
\quad- rope: cube, sphere, pulley, rope, fixed point\\
    \quad- fluid: fluid, container, stick, emitter\\
    \quad- cloth: cloth, pillar, plate\\
    \quad- ball: ball, stick, pit\\
\\
Examples: \texttt{EXAMPLE\_INSERT\_HERE}\\
\\
Query: \texttt{QUERY\_INSERT\_HERE}\\
\\
 \bottomrule
\end{tabular}
\vskip -0.1in
\end{table*}

\begin{figure*}[t]
\begin{lstlisting}[language=Python, xleftmargin=.03\textwidth, xrightmargin=.03\textwidth,firstnumber=1, caption=Qualitative example of a property question in the rope scene., label={eg1}]
# Is the mass of the green cube greater than half that of the purple sphere?

def execute_command(video):
    rope_scene = SoftScene(video, 'rope')
    green_cube = rope_scene.find("green cube")
    purple_sphere = rope_scene.find("purple sphere")
    
    green_cube_mass = rope_scene.query_single('mass', green_cube[0])
    purple_sphere_mass = rope_scene.query_single('mass', purple_sphere[0])
    
    if green_cube_mass is not None and purple_sphere_mass is not None:
        return bool_to_yesno(green_cube_mass > 0.5 * purple_sphere_mass)
    else:
        return "can not answer"
\end{lstlisting}
\end{figure*}

\begin{figure*}[t]
\begin{lstlisting}[language=Python, xleftmargin=.03\textwidth, xrightmargin=.03\textwidth,firstnumber=1, caption=Qualitative example of a goal-driven question in the fluid scene., label={eg2}]
# What can we do to let most of the pink fluid enter black container? | Remove green stick

def execute_command(video):
    fluid_scene = SoftScene(video, 'fluid')
    pink_fluid = fluid_scene.find("pink fluid")
    black_container = fluid_scene.find("black container")

    gd_init_event = fluid_scene.register_event([], "remove", "green stick")
    fluid_gd_scene = fluid_scene.init_dyn_simulation(gd_init_event)

    flag = fluid_gd_scene.happen([pink_fluid], "entering", [black_container])
    
    return bool_to_yesno(flag)
\end{lstlisting}
\end{figure*}

\begin{figure*}[t]
\begin{lstlisting}[language=Python, xleftmargin=.03\textwidth, xrightmargin=.03\textwidth,firstnumber=1, caption=Qualitative example of a predictive question in the cloth scene., label={eg3}]
# Is the blue pillar finally in touch with the yellow plate?

def execute_command(video):
    cloth_scene = SoftScene(video, 'cloth')
    blue_pillar = cloth_scene.find("blue pillar")
    yellow_plate = cloth_scene.find("yellow plate")
    
    pred_init_event = cloth_scene.register_event([], "simulate", "")
    cloth_pred_scene = cloth_scene.init_dyn_simulation(pred_init_event)
    
    flag = cloth_pred_scene.happen([blue_pillar, yellow_plate], "touching", "")
    
    return bool_to_yesno(flag)
\end{lstlisting}
\end{figure*}

\begin{figure*}[t]
\begin{lstlisting}[language=Python, xleftmargin=.03\textwidth, xrightmargin=.03\textwidth,firstnumber=1, caption=Qualitative example of a counterfactual question in the ball scene., label={eg4}]
# If we removed the red floating wall and other balls, which pit would the black ball drop into?

def execute_command(video):
    ball_scene = SoftScene(video, 'ball')
    black_ball = ball_scene.find("black ball")
    pits = ball_scene.find("pit")

    cf_init_event = ball_scene.register_event([], "remove", "red floating wall and other balls")
    ball_cf_scene = ball_scene.init_dyn_simulation(cf_init_event)

    for pit in pits:
        if ball_cf_scene.happen([black_ball], "droping", pit):
            return pit

    return "can not answer"
\end{lstlisting}
\end{figure*}

\begin{figure*}[b]
\begin{lstlisting}[language=Python, xleftmargin=.03\textwidth, xrightmargin=.03\textwidth]

class SoftScene:
    """A Python class representing one of a soft scene(rope/fluid/cloth/ball) and objects in the scene, as well as relevant information.

    Attributes
    ----------
    scene_name : str
        The name of the scene. rope/fluid/cloth/ball
    simulation: SoftSimulation
        A SoftSimulation object representing the simulation of the scene.
    video : torch.Tensor
        A tensor of the original video.
    frm_num : int
        The number of frames in the video.
    all_event_actions : list
        A list of all actions that can be taken in the scene.
    mode: str
        Online or offline. Online means the video information and dynamic scene information are simulated real-time. Offline means the video information is pre-simulated and stored in the disk.
    vid: str
        The video id of the video. Used in offline mode.

    Methods
    -------
    find(object_name: str)->List[SceneObject]
        Returns a list of SceneObject objects matching object_name with properties if any are found.
    query_pair(property: str, object1: SceneObject, object2: SceneObject)->Tuple(Union[float, int, None], Union[float, int, None])
        Return a tuple of the property values of two compared objects when comparing. If the property is not comparable or can not be queried, return Tuple(None, None).
    query_single(property: str, object: SceneObject)->Union[float, int, str, None]
        Return the property values of the object. If the object does not have the property, return None.
    register_event(scene_objects: list, action: str, attribute: str)->SceneEvent
        Create an event in the scene to initiate a simulation of counterfactual scene or predictive scene, based on the action and attribute. Return a SceneEvent object.
    init_dyn_simulation(dyn_init_event: SceneEvent)->SoftDynamicScene
        Init the simulation for dynamic scene, including predictive scene, counterfactual scene and goal-driven scene. Use dyn_init_event to simulate the dynamic scene.
    """
   def __init__(self, video: torch.Tensor, scene_name: str, start_frame: int = 0, end_frame: int = -1, mode: str = 'online', video_id: str = None):
        """Initializes a SoftScene object by the video and the scene_name. The scene_name is used to specify the scene type and initialize the simulation.

        Parameters
        ----------
        video : torch.Tensor
            A tensor of the original video.
        scene_name : str
            The name of the scene. rope/fluid/cloth/ball
        start_frame : int
            The start frame of the video. Default is 0.
        end_frame : int
            The end frame of the video. Default is -1.
        mode: str
            Online or offline. 
        video_id: str
            The video id of the video. Used in offline mode.
        """
        
        self.scene_name = scene_name
        self.video = video[start_frame:end_frame]
        self.frm_num = self.video.shape[0]
        self.vid = video_id
        self.mode = mode
        
        if self.mode == 'offline':
            assert self.vid is not None
        if self.mode == 'online':
            assert self.vid is None

        mrcnn_ann = forward_mrcnn(
            scene = self.scene_name, 
            input = self.video,
            input_type = 'video'
            )
            
\end{lstlisting}
\end{figure*}

\begin{figure*}[tp]
\begin{lstlisting}[language=Python, xleftmargin=.03\textwidth, xrightmargin=.03\textwidth,firstnumber=71]
        self.simulation = initialize_simulation(
            scene = self.scene_name,
            pred_ann = mrcnn_ann, 
            frame_num = self.frm_num,
            gt_flag = False, 
            )
        
        self.all_event_actions = ['increse','decrease','emit','remove','simulate']


    def find(self, object_name: str) -> list[SceneObject]:
        """Returns a list of SceneObject objects matching object_name with properties if any are found.
        Otherwise, returns an empty list.

        Parameters
        ----------
        object_name : str
            the name of the object to be found

        Returns
        -------
        List[SceneObject]
            A list of SceneObject objects matching object_name with properties.

        Examples
        --------
        >>> # return the red solid pulley
        >>> def execute_command(video) -> List[SceneObject]:
        >>>     rope_scene = SoftScene(video, 'rope')
        >>>     red_solid_pulley = rope_scene.find("red solid pulley")
        >>>     return red_solid_pulley

        >>> # How many blue objects are there in the video?
        >>> def execute_command(video) -> int:
        >>>     rope_scene = SoftScene(video, 'rope')
        >>>     blue_objects = rope_scene.find("blue")
        >>>     return len(blue_objects)
        """

        all_objects = self.simulation.find_all_objs()

        name_list = parse_name(object_name).split(' ')
        obj_feats = {
            'color': parse_color(name_list),
            'shape': parse_shape(name_list),
            'dynamic': parse_dynamic(name_list),
            'type': parse_type(name_list),
        }

        for k, v in obj_feats.items():
            if v is not None:
                object_candidates = [obj for obj in all_objects if getattr(obj, k) == v]

        return object_candidates


    def query_single(self, property: str, object: SceneObject) -> Union[float, int, str, None]:
        """Return the basic property value of the object. If the object does not have the property, return None. 
        Call query_pair instead of this function twice if comparing two objects.

        Parameters
        -------
        property : str
            A string describing the property to be queried.
        object: SceneObject
            The object to be queried.

        Returns
        -------
        Union[float, int, str, None]
            The property value of the object. If the object does not have the property, return None.
        """

        return self.query_pair(property, object, [None])[0]

    
\end{lstlisting}
\end{figure*}

\begin{figure*}[tp]
\begin{lstlisting}[language=Python, xleftmargin=.03\textwidth, xrightmargin=.03\textwidth,firstnumber=147]
    def query_pair(self, property: str, object1: SceneObject, object2: SceneObject) -> tuple(Union[float, int, None], Union[float, int, None]):
        """Return a tuple of the property values of two compared objects when comparing. If the property is not comparable or can not be queried, return Tuple(None, None). 
        This function is used to return the values to be compared. 

        Parameters
        -------
        property : str
            A string describing the property to be queried.
        object1: SceneObject
            First object in the comparison.
        object2: SceneObject
            Second object in the comparison.

        Returns
        -------
        Tuple(Union[float, int, None], Union[float, int, None])
            A tuple of the property values of two compared objects when comparing. If the property is not comparable or can not be queried, return Tuple(None, None).

        Examples
        --------
        >>> # Is the fluid density of the blue fluid larger than that of the red fluid?
        >>> def execute_command(video) -> str:
        >>>     fluid_scene = SoftScene(video, 'fluid')
        >>>     blue_fluid = fluid_scene.find("blue fluid")
        >>>     red_fluid = fluid_scene.find("red fluid")
        >>>     blue_fluid_density, red_fluid_density = fluid_scene.query_pair('density', blue_fluid, red_fluid)
        >>>     if blue_fluid_density is not None and red_fluid_density is not None:
        >>>         return bool_to_yesno(blue_fluid_density > red_fluid_density)
        >>>     else:
        >>>         return 'can not answer'

        >>> # Is the mass of the sphere greater than half that of the black cube?
        >>> def execute_command(video) -> str:
        >>>     rope_scene = SoftScene(video, 'rope')
        >>>     sphere = rope_scene.find("sphere")
        >>>     black_cube = rope_scene.find("black cube")
        >>>     sphere_mass, black_cube_mass = query_both('mass', sphere, black_cube)
        >>>     if sphere_mass is not None and black_cube_mass is not None:
        >>>         return bool_to_yesno(sphere_mass > 0.5*black_cube_mass)
        >>>     else:
        >>>         return "can not answer"
        """

        property_dict = {
            'mass': query_pair_mass,
            'tension': query_pair_tension,
            'density': query_pair_density,
            'elasticity': query_pair_elasticity,
            'plasticity': query_pair_plasticity,
            'bending': query_pair_bending,
        }

        if property in property_dict:
            return property_dict[property](object1[0], object2[0], scene=self.scene_name)
        else:
            raise Exception(f'Property {property} not supported.')


    def init_dyn_simulation(self, dyn_init_event: SceneEvent):
        """Init the simulation for dynamic scene, including predictive scene, counterfactual scene and goal-driven scene. Use dyn_init_event to simulate the dynamic scene.

        Parameters
        -------
        dyn_init_event : SceneEvent
            A SceneEvent object that describes an event to initiate the dynamic scene.

        Returns
        -------
        SoftDynamicScene
            A SoftDynamicScene object representing the dynamic scene simulation.
        """

        return SoftDynamicScene(self, dyn_init_event, mode=self.mode)


\end{lstlisting}
\end{figure*}

\begin{figure*}[tp]
\begin{lstlisting}[language=Python, xleftmargin=.03\textwidth, xrightmargin=.03\textwidth,firstnumber=222]
    def register_event(self, scene_objects: list, action: str, attribute: str) -> SceneEvent:
        """Create an event in the scene to initiate a simulation of counterfactual scene or predictive scene, based on the action and attribute. Return a SceneEvent object.

        Parameters
        -------
        scene_objects : list
            A list of objects in the scene. If the action is 'simulate' or 'remove', the scene_objects can be empty.
        action : str
            A verb of action describing the action to be taken. For example, 'remove', 'increase', 'decrease', 'emit', and 'simulate'
        attribute : str
            A noun of attribute describing the type of event. For example, 'mass' (for 'increase' or 'decrease'), 'fluid' (for 'emit') and '' (for 'simulate')
            It can also be '' if the event is enough to describe. 

        Returns
        -------
        SceneEvent
            A SceneEvent object representing the event.

        Examples
        --------
        >>> # If the green stick were removed, which stick would blue fluid pass? 
        >>> ...
        >>> cf_init_event = fluid_scene.event(green_stick, "remove", "stick")        
        >>> fluid_cf_scene = fluid_scene.init_dyn_simulation(cf_init_event)
        >>> ...

        >>> # If we want the green cube to move down, what can we do? | Increase the mass of the blue sphere 
        >>> ...
        >>> gd_init_event = rope_scene.event(blue_sphere, "increase", "mass")
        >>> rope_gd_scene = rope_scene.init_dyn_simulation(gd_init_event)
        >>> ...

        >>> # Does the green plate fall over?
        >>> ...
        >>> pred_init_event = rope_scene.event([], "simulate", "")
        >>> cloth_pred_scene = rope_scene.init_dyn_simulation(pred_init_event)
        >>> ...
        
        """
        
        if len(scene_objects) > 1:
            raise Exception('Only one object is supported now.')
        if len(scene_objects) == 0 and action != 'simulate':
            raise Exception('No object is supported now.')
        
        if len(scene_objects) == 0:
            obj = None
        else:
            obj = scene_objects[0]
        if action not in self.all_event_actions:
            raise Exception(f'Action {action} not supported.')
        
        if is_objects(attribute):
            object_names = parse_attribute(attribute)
            attribute = self.find(object_names)

        return create_event(obj, action, attribute)


\end{lstlisting}
\end{figure*}

\begin{figure*}[tp]
\begin{lstlisting}[language=Python, xleftmargin=.03\textwidth, xrightmargin=.03\textwidth,firstnumber=281]
class SoftDynamicScene:
    """A Python class representing the dynamic type of a soft scene(rope/fluid/cloth/ball) and objects in the scene, as well as relevant information. This class is based on a SoftScene. The objects in this scene is the same with the SoftScene, while the events and dynamics are different. This class is used for dynamic, including counterfactual scene simulation, predictive scene simulation, as well as goal-driven scene simulation.

    Attributes
    ----------
    scene : SoftScene
        The scene that this dynamic scene is based on.
    scene_name : str
        The name of the scene. The same with the scene.scene_name.
    init_event : SceneEvent
        The main event of the dynamic scene.
    simulation: SoftSimulation
        A SoftSimulation object representing the simulation of the scene. The same with scene.simulation.
    simulation_dyn: SoftSimulationDyn
        A SoftSimulationDyn object representing the dynamic simulation of the scene.
    mode: str
        Online or offline. Online means the video information and dynamic scene information are simulated real-time. Offline means the video information is pre-simulated and stored in the disk.
    vid: str
        The video id of the video. Used in offline mode.
    all_dyn_actions: list
        A list of all actions that can happen in the dynamic scene.
        
    Methods
    -------
    happen(scene_objects: list, action: str, target: str or list)->bool
        Check whether the action and target will happe in the dynamic scene. Return in the boolean format.

    """
    def __init__(self, scene: SoftScene, init_event: SceneEvent, mode='online'):
        """Initializes a SoftDynamicScene object by the scene and the init_event. The scene is used to specify the scene type and initialize the simulation. The init_event is used to initialize the dynamic simulation.
        """
        self.scene = scene
        self.init_event = init_event
        self.scene_name = scene.scene_name
        self.simulation = self.scene.simulation
        self.mode = mode
        self.vid = self.scene.vid

        if self.mode == 'offline':
            assert self.vid is not None
        if self.mode == 'online':
            assert self.vid is None
            
        self.simulation_dyn = initialize_dyn(
            scene=scene, 
            scene_name=scene.scene_name, 
            init_event=init_event, 
            mode=mode, 
            vid=self.vid
            )
            
        self.all_dyn_actions = ['entering', 'passing', 'motion', 'rotation', 'collision', 'falling', 'touching', 'droping']

    
\end{lstlisting}
\end{figure*}

\begin{figure*}[t]
\begin{lstlisting}[language=Python, xleftmargin=.03\textwidth, xrightmargin=.03\textwidth,firstnumber=335]
    def happen(self, scene_objects: list, action: str, target: str or list) -> bool:
        """Check whether the action and target will happe in the dynamic scene. Return in the boolean format.

        Parameters
        -------
        scene_objects : list
            A list of objects in the scene.
        action : str
            A noun describing the action that may happen. For example, 'entering', 'passing', 'motion', 'rotation', 'collision', 'falling', 'touching', 'droping'.
        target : str or list
            A noun describing the target of the action. For example, 'up' and 'down' for 'motion', 'clockwise' and 'anti-clockwise' for 'rotation', and 'red container' for 'entering'. The target can be ''. The target can also be a list containing SoftObjects.

        Returns
        -------
        bool
            True if the action will happen, otherwise False.

        Examples
        --------
        >>> # Is the green plate finally in touch with the gray pillar?
        >>> ...
        >>> green_plate = cloth_scene.find("green plate")
        >>> gray_pillar = cloth_scene.find("gray pillar")
        >>> pred_init_event = cloth_scene.event([], "simulate", "")
        >>> cloth_pred_scene = cloth_scene.init_dyn_simulation(pred_init_event)
        >>> flag = cloth_pred_scene.happen([green_plate], "touching", [gray_pillar])
        >>> ...

        >>> # Does the green plate fall over?
        >>> ...
        >>> green_plate = cloth_scene.find("green plate")
        >>> pred_init_event = cloth_scene.event([], "simulate", "")
        >>> cloth_pred_scene = cloth_scene.init_dyn_simulation(pred_init_event)
        >>> flag = cloth_pred_scene.happen([green_plate], "falling", "")
        >>> ...

        >>> # What can we do to make the pink ball drop into the right pit? | Remove the yellow floating wall and other balls
        >>> ...
        >>> pink_ball = ball_scene.find("pink ball")
        >>> right_pit = ball_scene.find("right pit")
        >>> gd_init_event = ball_scene.event([], "remove", "yellow floating wall and other balls")
        >>> ball_gd_scene = ball_scene.init_dyn_simulation(gd_init_event)
        >>> flag = ball_gd_scene.happen([pink_ball], "droping", [right pit])
        >>> ...

        >>> # If we removed the orange floating wall and other balls, which pit would the white ball drop into?
        >>> ...
        >>> white_ball = ball_scene.find("white ball")
        >>> pits = ball_scene.find("pit")
        >>> cf_init_event = ball_scene.event([], "remove", "orange floating wall and other balls")
        >>> ball_cf_scene = ball_scene.init_dyn_simulation(cf_init_event)
        >>> for p in pits:
        >>>     flag = ball_cf_scene.happen([white_ball], "droping", p)
        >>>     if flag:
        >>>         return p
        >>> return "can not answer"
        >>> ...
        """
        
        if len(scene_objects) > 1:
            raise Exception('Only one object is supported now.')
        
        obj = scene_objects[0]
        if action not in self.all_dyn_actions:
            raise Exception(f'Action {action} not supported.')
        
        if is_objects(target):
            object_names = parse_target(target)
            target = self.find(object_names)

        prediction = predict(obj, action, target)
        return prediction.happen()
\end{lstlisting}
\end{figure*}

\clearpage

\begin{figure*}[!t]
\begin{lstlisting}[language=Python, xleftmargin=.03\textwidth, xrightmargin=.03\textwidth,firstnumber=407, caption=Full API., label={listing}]


def bool_to_yesno(bool_answer: bool) -> str:
    """Returns a yes/no answer to a question based on the boolean value of bool_answer.

    Parameters
    ----------
    bool_answer : bool
        a boolean value

    Returns
    -------
    str
        a yes/no answer to a question based on the boolean value of bool_answer
    """
    return "yes" if bool_answer else "no"
\end{lstlisting}
\end{figure*}

\end{document}